\documentclass[conference,10pt]{IEEEtran}
\IEEEoverridecommandlockouts

\def\thesubsubsectiondis{\thesubsectiondis\arabic{subsubsection}}

\makeatletter
\def\subsubsection{%
  \@startsection
    {subsubsection}                 
    {3}                             
    {\parindent}                    
    {0.7ex plus 0ex minus 0ex}      
    {0.7ex plus .5ex minus 0ex}     
    {\normalfont\normalsize\itshape}
}
\makeatother

\usepackage{microtype}
\usepackage{graphicx}
\usepackage{subcaption}
\usepackage{caption}
\usepackage{booktabs} 

\graphicspath{{./images/}}
\DeclareGraphicsExtensions{.pdf,.jpeg,.png}

\usepackage{hyperref}
\usepackage{algorithm}
\usepackage{algorithmicx}
\usepackage[noend]{algpseudocode}

\usepackage{amsfonts}
\usepackage{amsmath}
\usepackage{amssymb}

\usepackage{cite}
\usepackage{textcomp}
\usepackage{xcolor}
\def\BibTeX{{\rm B\kern-.05em{\sc i\kern-.025em b}\kern-.08em
    T\kern-.1667em\lower.7ex\hbox{E}\kern-.125emX}}

\begin{document}
\raggedbottom
\setlength{\parskip}{3pt}

\IEEEpubid{\makebox[\columnwidth]{\hspace{0.4in} Accepted at ICDM 2020. \hfill} \hspace{\columnsep}\makebox[\columnwidth]{ }}

\title{Effects of Model Misspecification on Bayesian Bandits: Case Studies in UX Optimization}

\markboth{IEEE International Conference on Data Mining}{Effects of Model Misspecification on Bayesian Bandits}

\author{\IEEEauthorblockN{Mack Sweeney\IEEEauthorrefmark{1}\IEEEauthorrefmark{2},
Matthew van Adelsberg\IEEEauthorrefmark{1},
Kathryn Laskey\IEEEauthorrefmark{3}, and
Carlotta Domeniconi\IEEEauthorrefmark{2}}
\IEEEauthorblockA{\IEEEauthorrefmark{1}Capital One, McLean, VA, USA}
\IEEEauthorblockA{\IEEEauthorrefmark{2}Department of Computer Science, George Mason University, Fairfax, VA, USA}
\IEEEauthorblockA{\IEEEauthorrefmark{3}Department of Systems Engineering and Operations Research, George Mason University, Fairfax, VA, USA}
\IEEEauthorblockA{\{msweene2, klaskey, cdomenic\}@gmu.edu,\ matthew.vanadelsberg@capitalone.com}%
}

\maketitle

\begin{abstract}
Bayesian bandits using Thompson Sampling have seen increasing success in recent years.
Yet existing value models (of rewards) are misspecified on many real-world problem.
We demonstrate this on the User Experience Optimization (UXO) problem, providing a novel formulation as a restless, sleeping bandit with unobserved confounders plus optional stopping.
Our case studies show how common misspecifications can lead to sub-optimal rewards, and we provide model extensions to address these, along with a scientific model building process practitioners can adopt or adapt to solve their own unique problems.
To our knowledge, this is the first study showing the effects of overdispersion on bandit explore/exploit efficacy, tying the common notions of under- and over-confidence to over- and under-exploration, respectively.
We also present the first model to exploit cointegration in a restless bandit, demonstrating that finite regret and fast and consistent optional stopping are possible by moving beyond simpler windowing, discounting, and drift models.
\end{abstract}


\section{Introduction}\label{sec:intro}

Bayesian bandits using Thompson Sampling (TS) have enjoyed both applied and theoretical success in recent years~\cite{daniel_russo_tutorial_2018}. This approach is often chosen for adaptive design and analysis of digital experiments because, with an appropriately specified \textbf{value model} (of rewards), it provides a causal interpretation upon which optional stopping and winner selection rules can be built. These are critical for the human-in-the-loop optimization processes typically carried out via sequences of A/B tests. This interpretation is lacking in other state-of-the-art methods, which either apply heuristics to adapt simpler models or else employ black-box models~\cite{burtini_survey_2015}.

Despite its popularity, standard value models used for TS are often misspecified on real-world data~\cite{van_adelsberg_modeling_2019}.
Our aim is to clearly illustrate common data characteristics that result in biased inference from standard value models, making concrete the theoretical observations of~\cite{Phan_2019}, and to provide tools for identifying and handling them.
We do this through empirical case studies in the context of digital user experience (UX) optimization (UXO), simulating data to closely match those in industry experiments~\cite{van_adelsberg_modeling_2019}.
While UXO does not represent all domains, its challenges are common in other bandit applications.

We make the following contributions, all novel to our knowledge:

\begin{enumerate}
    \item Characterize UXO as a restless, sleeping bandit with unobserved confounders plus optional stopping.
    \item Demonstrate limitations of current Bayesian value models and provide alternatives to address them, along with a scientific model building process built on Bayesian model checking and nonstationary policy evaluation practitioners can adopt to solve their own unique problems.
    \item Empirically investigate the effects of overdispersion on bandit exploration/exploitation efficiency, tying the common notions of under- and over-confidence to over- and under-exploration, respectively.
    \item Develop the first model to exploit cointegration in a restless bandit, challenging Russo et al.'s assumption that "due to nonstationarity, no algorithm can promise regret that vanishes over time" for an important problem subclass~\cite{daniel_russo_tutorial_2018}.
\end{enumerate}

\subsection{Overview}

In Sections~\ref{sec:uxo-formalization}, \ref{sec:model-building-process}, and \ref{sec:models} we provide a literature review while formalizing the UXO problem, outlining our scientific model building process, and then describing standard Bayesian value models and our alternatives.
Section~\ref{sec:challenges} walks through several case studies, each diagnosing a problem with the value models considered up to that point and demonstrating an extended model that overcomes the problem, culminating in the cointegrated model.
We discuss the implications of these case studies in Section~\ref{sec:discussion}, outlining three strategies for overcoming the problems identified, both in UXO and beyond.
We then conclude in Section~\ref{sec:conclusion} and list directions for future work.

\section{User Experience Optimization (UXO)}\label{sec:uxo-formalization}

\subsection{Problem Overview}

\textbf{Goals:} UX designers often experiment with the design of websites, ads, and other UXs through sequential experimentation to increase visitor responses such as product purchases, ad clicks, customer acquisitions, or lifetime value.
Two or more UX variants are tested and a winner is selected then modified to generate the next variant(s) to test.
UXO aims to concurrently solve several subproblems to ultimately optimize some key performance indicator (KPI) of business value:
estimate the (1) sign and (2) magnitude of the gain/loss from a variant;
(3) if positive, make it permanent (rollout);
(4) if negative, rapidly control losses by routing visitors away from under-performing UXs; and
(5) explain the gain/loss estimate to inform future variant creation efforts.

\textbf{Economics:} The A/B/n test, designed and analyzed using null hypothesis significance testing (NHST), is the most common experimental method for this purpose~\cite{econsultancy_conversion_2017}.
NHST involves a static routing policy, i.e.\ A and B each get 50\% of test traffic -- which may be a subset of page traffic -- until a sufficient sample is obtained.
The false positive (FP) and false negative (FN) rate guarantees of NHST are important in domains like manufacturing, agriculture, or drug development, where creating variants and switching from the control are typically expensive.
In digital UXO, these costs are typically treated as negligible, reducing FP and FN costs to equivalent per-sample opportunity costs of suboptimal response rates, and the problem of minimizing this can be formulated as a restless, sleeping bandit with unobserved confounders, plus optional stopping.

\subsection{Preliminaries}

\textbf{Notation:} $N_t \in \mathbb{Z}^{*^A}$ indicates $N_t$ is the $t^\text{th}$ row in a matrix $N$ and an $A$-dimensional row vector with values in the set of non-negative integers. $X \in \mathbb{R}^{T \times A}$ indicates $X$ is a $T \times A$-dimensional matrix. Outside this context, capital letters without subscripts are constants.

\textbf{Data:} Feedback is delayed for many KPIs of interest.
For this and other reasons discussed below, many real-world applications are built on batch data pipelines.
Most experiments optimize for a binary reward metric, or else a real-valued metric that is a function of a binary metric.
So we limit ourselves to study of binary rewards.
At each time step $t$, the agent observes a batch of sample sizes $N_t \in {\mathbb{Z}^*}^A$ and response counts $Y_t \in {\mathbb{Z}^*}^A$ for each experience $a = 1, \ldots A$.
For concreteness and without loss of generality, we subsequently refer to time units as days.

\subsection{Problem Formalization}

\textbf{Bandit}: We'll begin our UXO formalization as a \textit{stochastic multi-armed bandit (MAB)} \cite{burtini_survey_2015}.
For each visitor $n = 1, \ldots, N$, the agent chooses an experience $a_n \in \{1, \ldots, A\}$ according to its policy $\pi: \{1, \ldots, N\} \to \{1, \ldots, A\}$.
The visitor then responds according to the unknown but learnable distribution of the associated rate, with mean $\theta_{a_n}$.
The agent's goal is to maximize the response count, $\sum_{n=1}^N y_n$, where $y_n$ is visitor $n$'s binary response.
Several aspects of UXO complicate this goal.

\begin{enumerate}
  \item  \textbf{Delayed feedback}~\cite{Chapelle_2014}: Visitors often do not respond immediately, necessitating an auxiliary attribution model, which we assume given but note is critical. Furthermore, many experimentation systems provide data via batch pipelines, which may also be delayed and occasionally broken. We proceed with a batch data setting, though we do assume batches arrive at a consistent time interval.
  \item  \textbf{Overdispersion}~\cite{barron_analysis_1992}: The KPI is often a non-click-through response such as product purchase, and visitors are often prospects, for which only cookie data is available.
  This can lead to unobserved heterogeneity in the population and/or nonstationarity.
  These are the two sources of overdispersion -- the situation where the data has more variance than the model can represent.
  This is a problem for the generalized linear models (GLMs) commonly used for modeling count-valued responses in Bayesian bandits.
  \item  \textbf{Restless -- Drifting}~\cite{whittle_restless_1988}: Time-dependent rates $\{\theta_a\}_{a=1}^A$ may exhibit seasonality, linear or non-linear trends, or some other repeating patterns.
  The preferred solution in these cases is to control for the covariates inducing time-dependence; however, not all temporal patterns are consistent or repeating, and the covariates inducing these more irregular patterns are often unobservable and suitable instrumental variables unavailable.
  For instance, gradual shifts in unobserved aspects of the visitor population mix may be driven by exogenous intra- or inter-company marketing.
  This sort of uncontrolled time-dependence often manifests as random walk or Brownian motion dynamics, which we refer to generally as ``drifting rates."
  Such drift dynamics are particularly likely when responses involve a financial outlay, tying them to macroeconomic trends~\cite{nelson_trends_1982}.
  \item  \textbf{Sleeping}~\cite{kleinberg_regret_2010}: UX designers typically want to rapidly test a sequence of hypotheses, rather than optimally allocating between a fixed arm set.
  Arms are removed and new arms added by iterating on winners.
\end{enumerate}

\textbf{Optional Stopping}~\cite{de_heide_why_2017}: In practice, the agent controls arm availability, which differs from the standard sleeping bandit.
This task is critical in UXO.
Thus stopping rule selection and evaluation are essential solution elements.
While often implicitly considered part of the bandit~\cite{scott_multi-armed_2015}, recognizing it as the optional stopping problem makes clear the importance of stopping rule selection and evaluation.

\subsection{Excluded Problem Aspects}

There are also important aspects of UXO that we exclude from this study.

\begin{enumerate}
  \item  \textbf{Restless -- Changepoints}~\cite{Luo_Agarwal_Langford_2018}: Response rates affected by abrupt changes also fall into the restless bandit subclass. Such changepoints may arise due to temporary system outages or data quality issues, in which case they tend to show up as temporary spikes or dips in response rates. They may also arise due to changes in typical visitor pathways through the digital product being tested, e.g. the introduction of a new link redirects many visitors into or away from the UX being optimized. This latter type of changepoint usually shows up as a step change in response rates that persists.
  \item  \textbf{Mortal-Armed and Rotting}~\cite{chakrabarti_mortal_2009, levine_rotting_2017}: A related but separate category of problems are the mortal-armed and rotting bandit.
  In the former, some arms become unavailable when a fixed budget is consumed or a fixed time period elapses.
  In the latter, the expected reward of each arm decays as a function of the number of pulls.
  In UXO, the agent decides when an arm dies, rather than a fixed budget or time period controlling that outcome.
  While some UXO such as marketing-type content placement may be subject to reward decay, others such as basic usability improvements are not.
  \item  \textbf{Structured}~\cite{lattimore_bounded_2014, gupta_unified_2019}: UXO also frequently involves multivariate analysis of correlated, factorized treatments and population segmentation resulting in a contextual bandit.
  Both can be represented as variations of structured bandits with optional stopping.
\end{enumerate}
\section{Scientific Model Building Process}\label{sec:model-building-process}

\subsection{Agent Construction}

The whole model, or agent, consists of a Bayesian value model, an adaptive design (bandit) policy, and an optional stopping rule.
As responses arrive, it uses these three components to (1) update the posterior distribution for the response rate of each UX, (2) optionally stop, otherwise (3) devise a policy used until the next update.
The value model, discussed in detail in the next section, is our focus.
So the policy and basic stopping rule are fixed as follows.

\textbf{Policy}: The most common Bayesian policy-making methods are (1) variants of Thompson Sampling (TS), which match routing proportions to the probability each experience is best, and (2) variants of Upper Confidence Bound (UCB), which treat some large quantile as an optimistic reward estimate and select the experience with the highest UCB~\cite{daniel_russo_tutorial_2018}.
Since randomized policies have been shown more resilient to feedback delay~\cite{chapelle_empirical_2011}, we focus exclusively on TS.

\textbf{Stopping Rule}: As our case studies demonstrate, this is often a composition of rules/heuristics.
Our basic rule is a common choice: declare any arm the winner once it has at least 95\% probability to be best~\cite{scott_multi-armed_2015}.
There are a variety of alternatives in the literature on best arm identification (BAI)~\cite{audibert_best_2010, russo_simple_2016}, racing algorithms~\cite{maron_racing_1997, even-dar_action_2006}, satisficing~\cite{russo_time-sensitive_2017}, and the $(\epsilon, \delta)\text{-PAC}$ framework~\cite{garivier_non-asymptotic_2019}.
We leave study of these to future work, noting that a better value model generally maps to better stopping outcomes regardless of the rule using it.

\subsection{Evaluation}

We consider three diagnostic procedures to assess how well an agent maximizes the KPI of interest.
\textbf{Nonstationary policy evaluation (NPE)} measures agent performance in KPI units -- necessary for assessing real utility of improvements -- but is slow and expensive.
\textbf{Posterior parameter visualization} and \textbf{posterior predictive checks (PPCs)} measure value model goodness of fit with other metrics.
An improved value model equips an agent to increase the KPI.
So we recommend iterating on value model fitness with these methods, then comparing agent performance with NPE.
In our case studies, our NPE takes into account agent stopping decisions, but we still update its posterior and policy for the purposes of visualization and PPCs.

Visualizing parameter estimates is a standard method of diagnosing obvious mis-estimation issues.
PPCs and NPE require some additional explanation.

\subsubsection{PPCs: Value Model Fit Quality}\label{subsubsec:ppcs}

The posterior predictive (PP) distribution (PPD) represents our beliefs about unobserved data given our model and observations.
Formally, it is obtained by marginalizing over our posterior uncertainty, but in practice it is typically estimated via MC.
Let the set of all parameters be defined as $\Theta$ and a MC sample $s$ from the joint posterior be $\Theta^{(s)}$, then:
\vskip -0.2in
\begingroup
\allowdisplaybreaks
\begin{align*}
    Y^{pred} &\sim p(Y^{new} | N^{new}, Y, N) \\
        \noindent &= \int p(Y^{new} | N^{new}, \Theta) p(\Theta | Y, N) \partial\Theta \\
        &\approx \frac{1}{S} \sum_{s=1}^S p(Y^{new} | N^{new}, \Theta^{(s)})
\end{align*}
\endgroup

\noindent PPCs compare PPD statistics to their corresponding empirical quantities to evaluate goodness-of-fit and are standard for Bayesian model checking~\cite{bda3}.
To diagnose issues with our binary response models, we compare statistics of the PP rates $\hat{\theta} = Y^{pred} / N$ to those of the empirical rates $\bar{\theta} = Y / N$.

\textbf{Coverage} is a versatile PPC for overall probabilistic calibration~\cite{Gneiting_Balabdaoui_Raftery_2007}.
All our models produce a distribution over $\hat{\theta}$ for each UX for each of the $T$ days.
Coverage is then computed as the proportion of the $T$ $\bar{\theta}_{*,a}$ within their respective width-$W$ ($\in (0, 1)$) credible intervals (CIs).
For a well-calibrated model, this will be close enough to $W$ for the difference to be explained by sampling variability.~\footnote{E.g., with 90\% CIs and $T=10$ days, about 9 of 10 daily empirical averages should be within their respective CIs, but $T$ is small enough for values of 7, 8, and 10 to have reasonably come from a well-calibrated model.}
So we compare the true coverage to a distribution of plausible coverages assuming perfect calibration: $Binomial(n=T, p=W)$.
This provides a simplistic but useful PP $p$-value.~\footnote{This assumes independence between the per-day within-CI indicator variables, which is likely false. Hence, the resulting $p$-values should be interpreted with some caution.}
We also include coverage plots (e.g. Fig.~\ref{fig:fixed-bb-coverage}), since patterns in which points are over or under the CI help diagnose temporal bias~\cite{Gelman_Shalizi_2013}.

\subsubsection{NPE: Holistic Agent Performance}\label{subsubsec:npe}

\textbf{Measurement}: While PPCs are useful for assessing model fit, they do not tell us how well the agent will perform in a given environment in the units of reward.
This question is counterfactual; unlike in supervised learning, we cannot simply fit on a training set and evaluate on a test set; the agent's policy determines the data it will observe.
The two approaches to NPE are: (1) simulate an environment which approximates the real one, or (2) use a counterfactual estimator on data logged from the same or a similar environment~\cite{dudik_sample-efficient_2012}.
To simplify the exposition of our case studies by ensuring each agent sees the same dataset, we take the latter approach, simulating a fixed dataset for each case study.

\textbf{Estimator}:
For our counterfactual estimator, we choose Dudik et al.'s \textbf{DR-ns} ("doubly robust nonstationary") estimator for its high sample-efficiency~\cite{dudik_sample-efficient_2012} and set $q = 0.01$ to accept a small amount of bias in order to reduce variance.
DR-ns uses rejection sampling: each day, the agent's ``nonstationary policy" determines what subset of the next day's data is observed.
So we conduct multiple repetitions for all experiments, with random seeds unique per (agent, repetition) pair.
The output from each repetition is an estimate of the reward rate the agent would have obtained on the dataset.

\textbf{Comparison}: To compare agent reward rates, we use a Bayesian hierarchical linear model with 2-way interactions to alleviate the multiple comparisons problem with minimal power loss~\cite{gelman_why_2012}.
Dummy-encoded categorical covariates are included for value model and stopping rule.
Each categorical's main effects are grouped under common priors and all interaction effects are grouped under one common prior (three total groups).
Normal-Inverse-Gamma($\mu{=}0,\alpha{=}2,\beta{=}\sigma$), with $\sigma \sim \text{Gamma}(\alpha{=}4,\beta{=}2)$.
The same model could be used to compare performance in simulators and, by including additional covariates for policy-making methods and datasets, to compare more diverse agents on and across datasets.

\section{Improved Bayesian Value Models for UXO}\label{sec:models}

We now present standard models and our extensions that are more successful in UXO. For all models, hyperparameters are set to standard uninformative options, or based on our intuition of the problem, rather than tuned. Except in cases of extreme data sparsity not represented in our case studies, results are not sensitive to these choices.

\subsection{Standard Bayesian Value Models}\label{subsec:standard-models}

\subsubsection{\textbf{IBB}}

Assuming the responses for each UX are drawn from Binomials with independent Beta-distributed rates is perhaps the simplest choice~\cite{scott_multi-armed_2015}. The Beta distribution is parameterized in terms of the mean and sample size throughout this paper~\cite{jorgensen_proper_1997}: 

\vskip -0.23in
\begin{equation*}
    \begin{aligned}
        Y_{t,a} &\sim Binomial(N_{t,a}, \theta_a)
    \end{aligned}
    \quad
    \begin{aligned}
        \theta_a &\sim Beta(\mu_a, \eta_a)
    \end{aligned}
\end{equation*}
\vspace{-0.22in}

\subsubsection{\textbf{Logistic}}

Replacing the Beta priors with logit-Normals gives a Logistic Regression (LR) with independent experience effects:

\vskip -0.15in
\begin{equation*}
    \begin{aligned}
        \theta_a &= \text{logit}^{-1}(\beta_a)
    \end{aligned}
    \quad
    \begin{aligned}
        \beta_a &\sim \mathcal{N}(\mu_a, \sigma_a^2)
    \end{aligned}
\end{equation*}
\vspace{-0.2in}

\noindent Use of LR for bandits was mentioned by Scott~\cite{scott_multi-armed_2015} and studied in detail by Dumitrascu et al.~\cite{dumitrascu_pg-ts_2018}. To improve regularization via shrinkage, we consider the hierarchical extension, where the experience effects are assumed drawn from a shared prior:

\vskip -0.18in
\begin{equation*}
    \begin{aligned}
        \beta_a &\sim \mathcal{N}(\mu, \sigma^2)  \\
        \mu &\sim \mathcal{N}(\mu_0, 1)
    \end{aligned}
    \quad
    \begin{aligned}
        \sigma^2 &\sim InvGamma(\nu/2, \nu/2), \nu = 5  \\
        \mu_0 &= \text{logit}(\bar{\theta}_{t,a})
    \end{aligned}
\end{equation*}
\vspace{-0.12in}

\noindent The prior mean $\mu_0$ is set to an empirical Bayes estimate -- the logit of the empirical cumulative response rate -- and the prior on the coefficients is equivalent to a $Student's{-}t(\mu, \sigma, \nu)$, with degrees of freedom $\nu$ chosen per guidance from~\cite{ghosh_use_2018}.

\subsection{Challenges and Extensions}

Our case studies in the next section illustrate the following problems, addressed by the models laid out in this section.%
\begin{enumerate}
    \item  Overdispersion due to unobserved heterogeneity causes early under-exploration, delaying stopping.
    \item  Under-exploration when rates are drifting biases data collection, preventing convergence to the true winner.
    \item  Ignorance of cointegration causes over-exploration, delaying or preventing stopping.
    \item  When introducing new variants and using previous data, estimation is biased towards/against them if recent drift has been upwards/downwards.
\end{enumerate}

\subsubsection{Overdispersion -- \textbf{BB-GLM}}
As our first case study shows (Section~\ref{subsec:heterogeneity}), overdispersion leads to over-confidence, which causes under-exploration.
If the rates are stationary, this can manifest as bias towards arms with lucky streaks early on but will eventually, if slowly, correct itself.
If the rates are nonstationary, as in the second case study (Section~\ref{subsec:drifting}), the bias may persist permanently.
We model overdispersion by assuming the population viewing each UX varies across days, resulting in a daily Binomial rate $\mu_{t,a}$ which we assign a Beta prior (for similar models, see~\cite{cepeda-cuervo_double_2017}).
The Beta means are the fixed response rates and their sample sizes are functions of dispersion parameters $\gamma_a$, which are pooled towards a common prior:

\begingroup
\allowdisplaybreaks
\vskip -0.15in
\begin{equation*}
    \begin{aligned}
        Y_{t,a} &\sim Binomial(N_{t,a}, \mu_{t,a})  \\
        \mu_{t,a} &\sim Beta(\theta_a, \eta_a)  \\
        \eta_a &= \gamma_a^{-1} - 1
    \end{aligned}
    \quad
    \begin{aligned}
        \gamma_a &\sim Beta(\lambda, \nu)  \\
        \nu^{-1} &\sim \mathcal{N}(0, 0.1)  \\
        \lambda &\sim Beta(\lambda_0, \nu_0)
    \end{aligned}
\end{equation*}
\endgroup

\noindent $\nu_0 = 7 \times |A|$ and $\lambda_0$ is set s.t.\ $p(\mu_{1,a})$ has a 90\% CI width of 0.8 using the Nelder-Mead Simplex algorithm.
These hyperparameter choices infuse skepticism during the first 7 days while avoiding subsequent over-reliance on the prior.
$\gamma_a \in (0, 1)$ is the degree of overdispersion, i.e.\ extra variation in the population observing $a$.
$\text{Var}(\mu_{t,a}) \rightarrow 0$ as $\gamma_a \rightarrow 0$; a theoretical $\gamma_a = 0$ gives a point mass on $\theta_a$ (no overdispersion) and 1 a uniform distribution.

\subsubsection{Drift -- \textbf{BB-Drift}}
As can be seen in our second case study (Section~\ref{subsec:drifting}), BB-GLM can mitigate bias when rates are drifting, but it comes at the cost of exploitation-stunting uncertainty that prevents or delays winner selection.
Common heuristic solutions force a fixed amount of exploration, e.g.\ reserve 10\% of traffic for equal allocation, or discount/window older data~\cite{raj_taming_2017}.
The former can help ensure more sufficient randomization across time at a fixed opportunity cost, while the latter can reduce the impact of biased data collection, but both require tuning.
For example, with too little windowing/discounting, the agent won't mitigate bias effectively, but with too much, it will discard so much data that confidence in a winner cannot be sustained.

The limitations of heuristic tuning are avoided with a model-based alternative.
Motivated by Granmo and Berg's incorporation of random walk dynamics in the coefficient estimates of per-arm linear models~\cite{granmo_solving_2010}, we explore a dynamic extension of BB-GLM:

\vskip -0.14in
\begingroup
\allowdisplaybreaks
\begin{equation*}
    \begin{aligned}[c]
        Y_{t,a} &\sim Binomial(N_{t,a}, \mu_{t,a})  \\
        \mu_{t,a} &\sim Beta(\theta_{t,a}, \eta_a)  \\
        \theta_{t,a} &= \text{logit}^{-1}(\beta_{t,a})
    \end{aligned}
    \quad
    \begin{aligned}[c]
        \beta_{1,a} &\sim \mathcal{N}(\mu, \sigma^2)  \\
        \beta_{t,a} &\sim \mathcal{N}(\beta_{t-1,a}, \rho_a^2)  \\
        \rho_a &\sim \mathcal{N}(0, \rho_0^2)
    \end{aligned}
\end{equation*}
\endgroup

\noindent $\eta_a, \mu, \sigma^2$ are specified as in BB-GLM.
$\rho_0^2$ is set to give a 90\% probability of daily rate changes at most $\delta$, which can be set using domain knowledge.
In our experiments, $\delta=0.005$.

\subsubsection{Cointegration -- \textbf{BB-Coint}}
As our second and third case studies demonstrate (Sections~\ref{subsubsec:cointegration} and \ref{subsubsec:cointegration}), drift models have a key limitation studied extensively in economics as a failure to capture \textbf{cointegration}, i.e.\ a stationary difference between nonstationary series~\cite{granger_introduction_1991}.
Without formal proof, we make the intuitive claim that \textit{a stable winner can exist in the limit of a nonstationary problem if there is cointegration}.
Restless bandit problems with cointegration thus represent an important but previously unstudied subclass for which vanishing regret is achievable.
We extend BB-Drift to capture cointegration as follows.
We extend BB-Drift to capture cointegration by assuming the effects of all non-control UXs are multiplicative offsets $\phi_a, a > 1$ from the control:

\vspace{-0.2in}
\begin{equation*}
    \begin{aligned}
        \beta_{1,1} &\sim \mathcal{N}(\mu_0, 1)  \\
        \beta_{t,1} &\sim \mathcal{N}(\beta_{t-1,1}, \rho^2)  \\
        \beta_{t,a} &= (1 + \phi_a \times sign(\beta_{t,1})) \times \beta_{t,1}
    \end{aligned}
    \quad
    \begin{aligned}
        \rho &\sim \mathcal{N}(0, \rho_0^2)  \\
        \phi_a &= \mathcal{N}(0, 1)
    \end{aligned}
\end{equation*}

\noindent The sign term ensures a positive $\phi_a$ shifts upwards and vice versa.
Hyperparameters are set as in previous models.

\subsection{Inference and Generalizations}

\textbf{Inference}: While the IBB posterior can be inferred using conjugacy, the rest must be approximated.
For this, we used Markov Chain Monte Carlo (MCMC) inference, specifically the No U-Turns Sampler (NUTS) implemented in Stan~\cite{stan}.
For simplicity and since TS uses Monte Carlo (MC) posterior draws, we also represent the IBB posterior as draws from its Beta posterior.
So inference for all models yields $S$ MC samples from the joint posterior of the response rates for each UX, $\hat{\theta} \in \mathbb{R}^{S \times A}$.

\textbf{Generalizations}: The problems studied here also arise in the structured variations of UXO and in similar bandit problems in other domains.
All our models are Dynamic GLMs~\cite{west_dynamic_1985} and thus can be extended to structured bandits.
Specifically, let $X_n \in \mathbb{R}^A$ be the covariates for visitor $n$, which only contains an indicator for which variant was seen.
Then linear predictor $X_n^T \beta_t = \beta_{t,a}$, which reduces to $\beta_a$ assuming static effects.
For cointegration, deterministic components can be added for additional control variables and the Error Correcting Model leveraged to automatically determine the cointegration vector for an arbitrary number of variants~\cite{koop_bayesian_2006}.
Finally, we note that while many practical applications require only a batch policy update, online versions of these models can be efficiently implemented via data augmentation, Gibbs sampling, and Kalman Filtering~\cite{polson_bayesian_2013,aktekin_sequential_2018,koop_bayesian_2006}.

\begin{figure*}[t]
    \centering
    \captionsetup[subfigure]{aboveskip=2pt,belowskip=2pt}
    \begin{subfigure}[t]{0.32\textwidth}
        \vskip 0pt
        \centering
        \includegraphics[width=\textwidth,height=2.5cm]{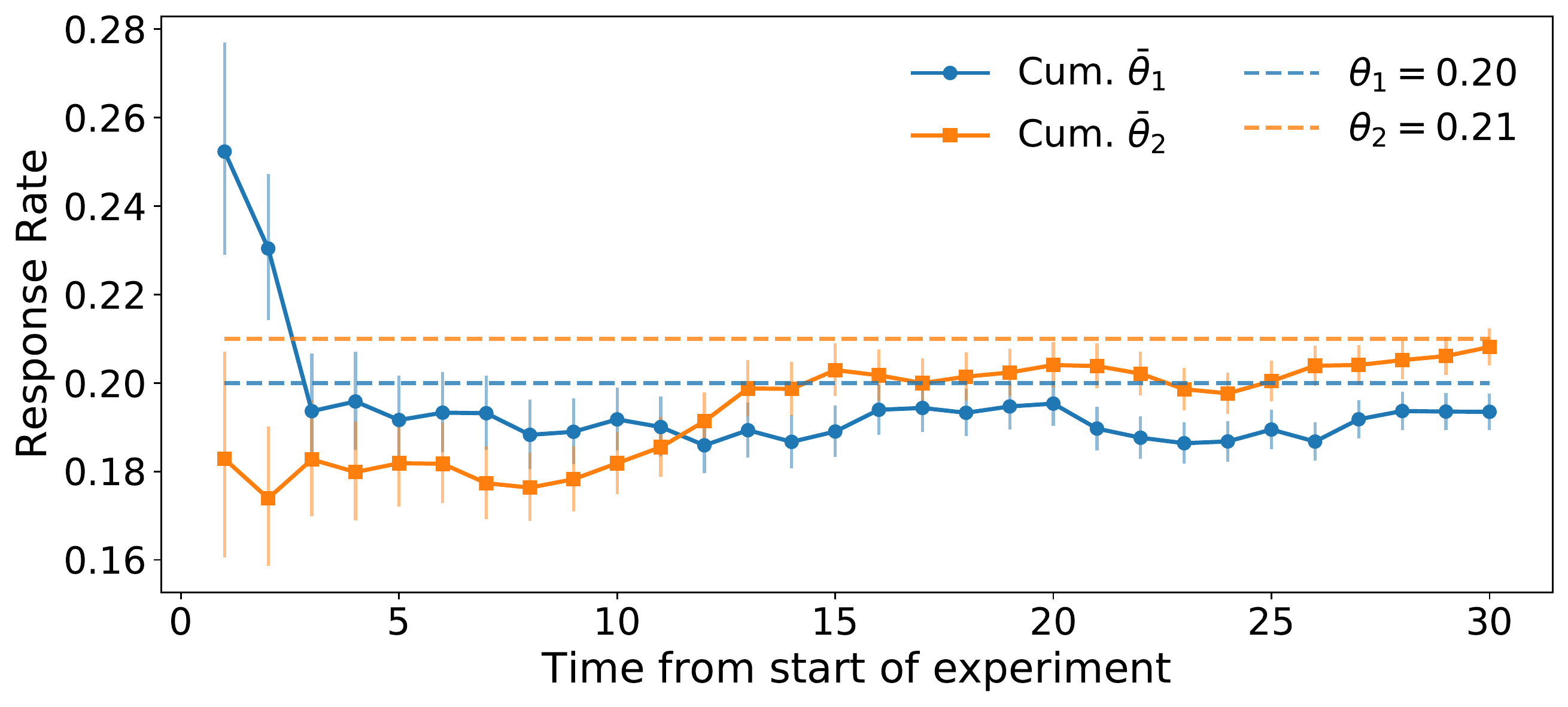}
        \caption{Dataset with fixed effects, overdispersion due to unobserved heterogeneity, 5\% lift, and daily arrival rates of 500 visitors each.
                 Error bars are 67\% CIs from a Beta-Binomial with a Jeffrey's prior.}
        \label{fig:fixed-data}
    \end{subfigure}%
    \hfill
    \begin{subfigure}[t]{0.32\textwidth}
        \vskip 0pt
        \centering
        \includegraphics[width=1.05\textwidth, height=3.15cm]{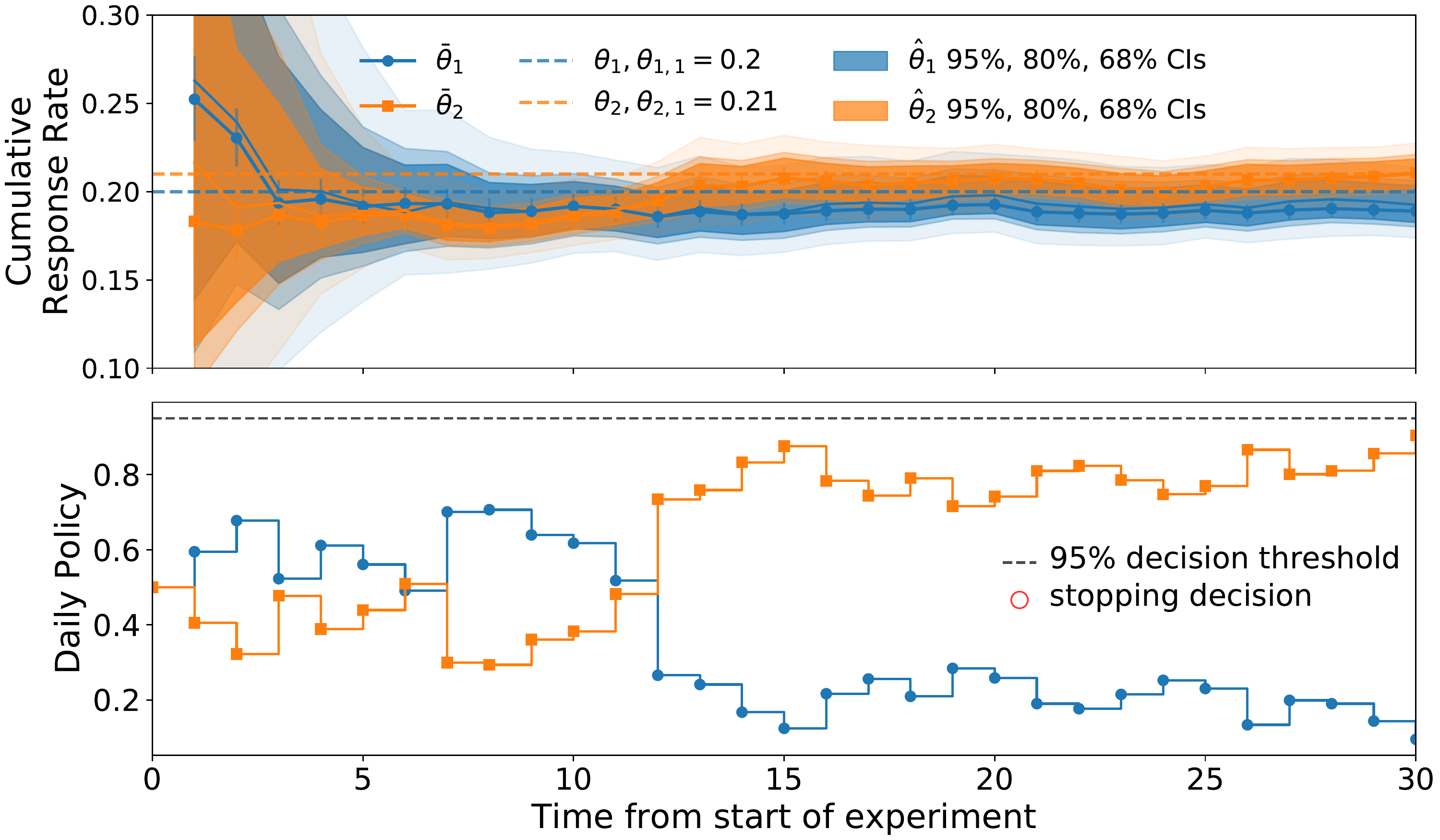}
        \caption{BB-GLM robustness to early noisy data prevents the overconfident exploitation seen in the Logistic and IBB.}
        \label{fig:fixed-bb}
    \end{subfigure}
    \hfill
    \begin{subfigure}[t]{0.32\textwidth}
        \vskip 0pt
        \centering
        \includegraphics[width=\textwidth,height=3.15cm]{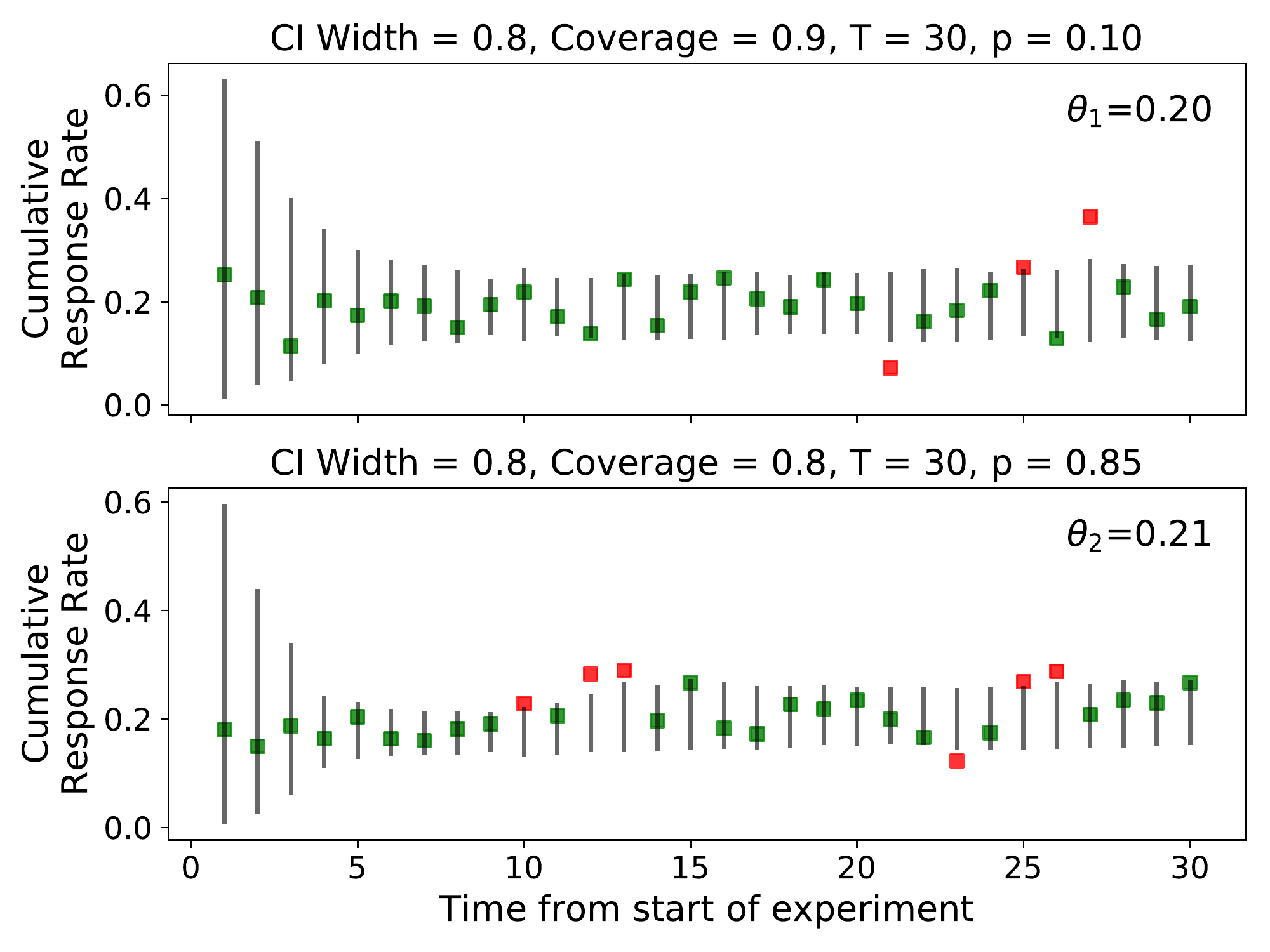}
        \caption{BB-GLM coverage shows good calibration. Vertical lines are the 80\% $\hat{\theta}_{t,a}$ CIs. Points are the $\bar{\theta}_{t,a}$, green if within the CI, else red.}
        \label{fig:fixed-bb-coverage}
    \end{subfigure}
    \hfill
    \begin{subfigure}[t]{0.32\textwidth}
        \vskip 0pt
        \centering
        \includegraphics[width=1.05\textwidth,height=3.15cm]{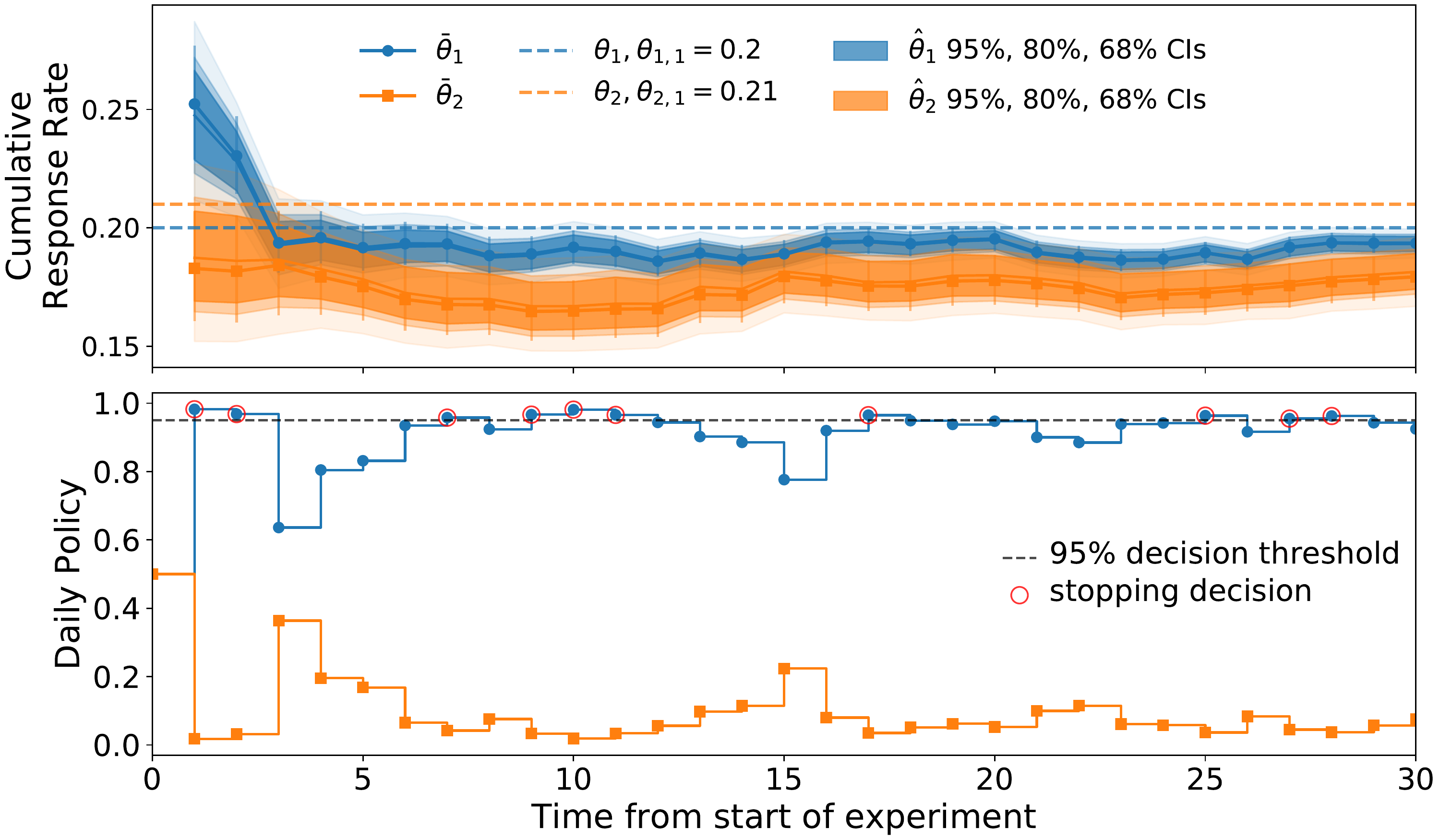}
        \caption{Logistic value estimate and policy plots indicate initial overconfidence biases data collection and worsens allocations.}
        \label{fig:fixed-logistic-bias}
    \end{subfigure}%
    \hfill
    \begin{subfigure}[t]{0.32\textwidth}
        \vskip 0pt
        \centering
        \includegraphics[width=1.05\textwidth,height=3.15cm]{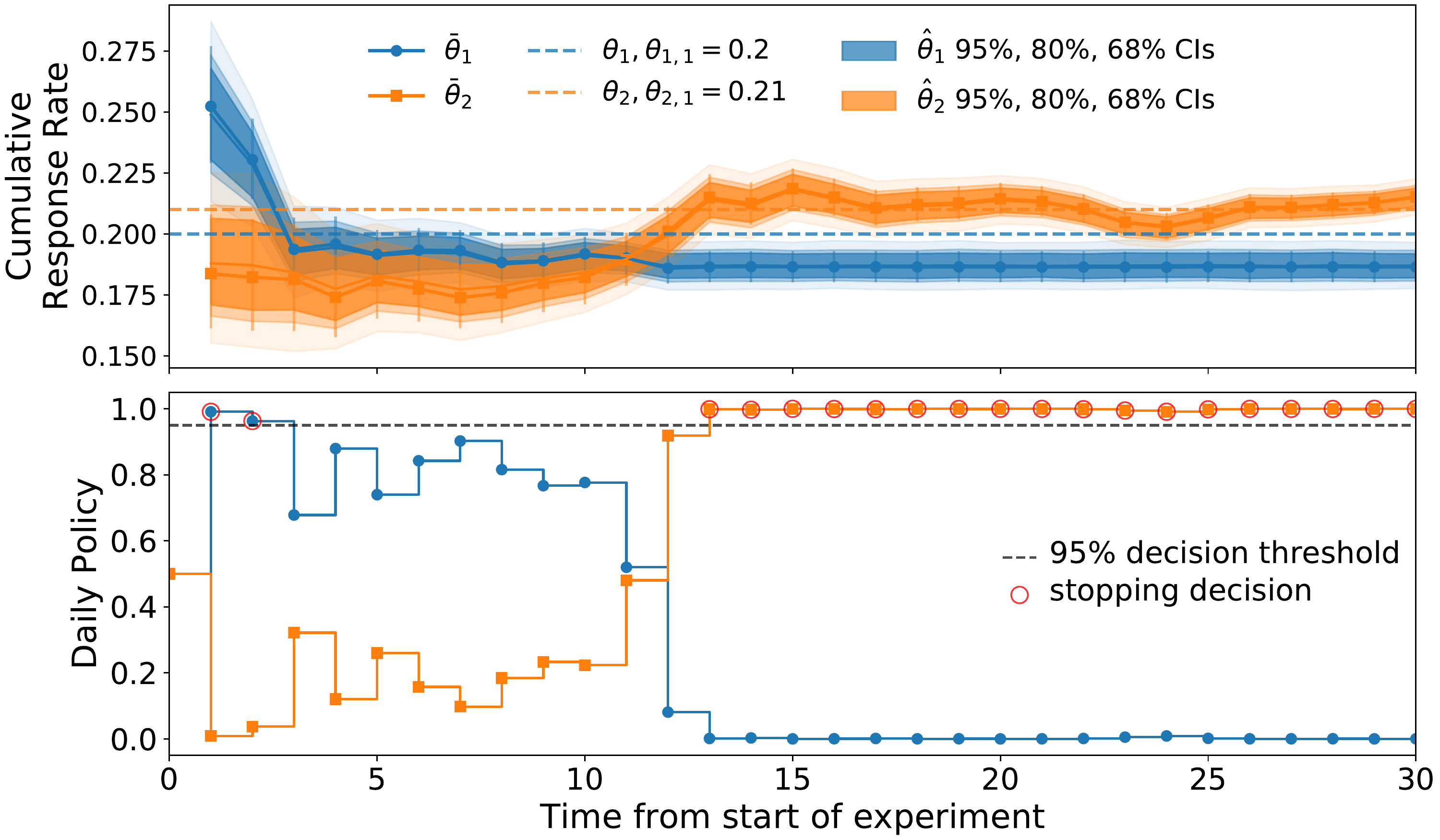}
        \caption{Logistic behavior on fixed but overdispersed dataset where initial overconfidence is corrected within the $T$ days.}
        \label{fig:fixed-logistic-corrected}
    \end{subfigure}
    \hfill
    \begin{subfigure}[t]{0.32\textwidth}
        \vskip 0pt
        \centering
        \includegraphics[width=\textwidth,height=3.15cm]{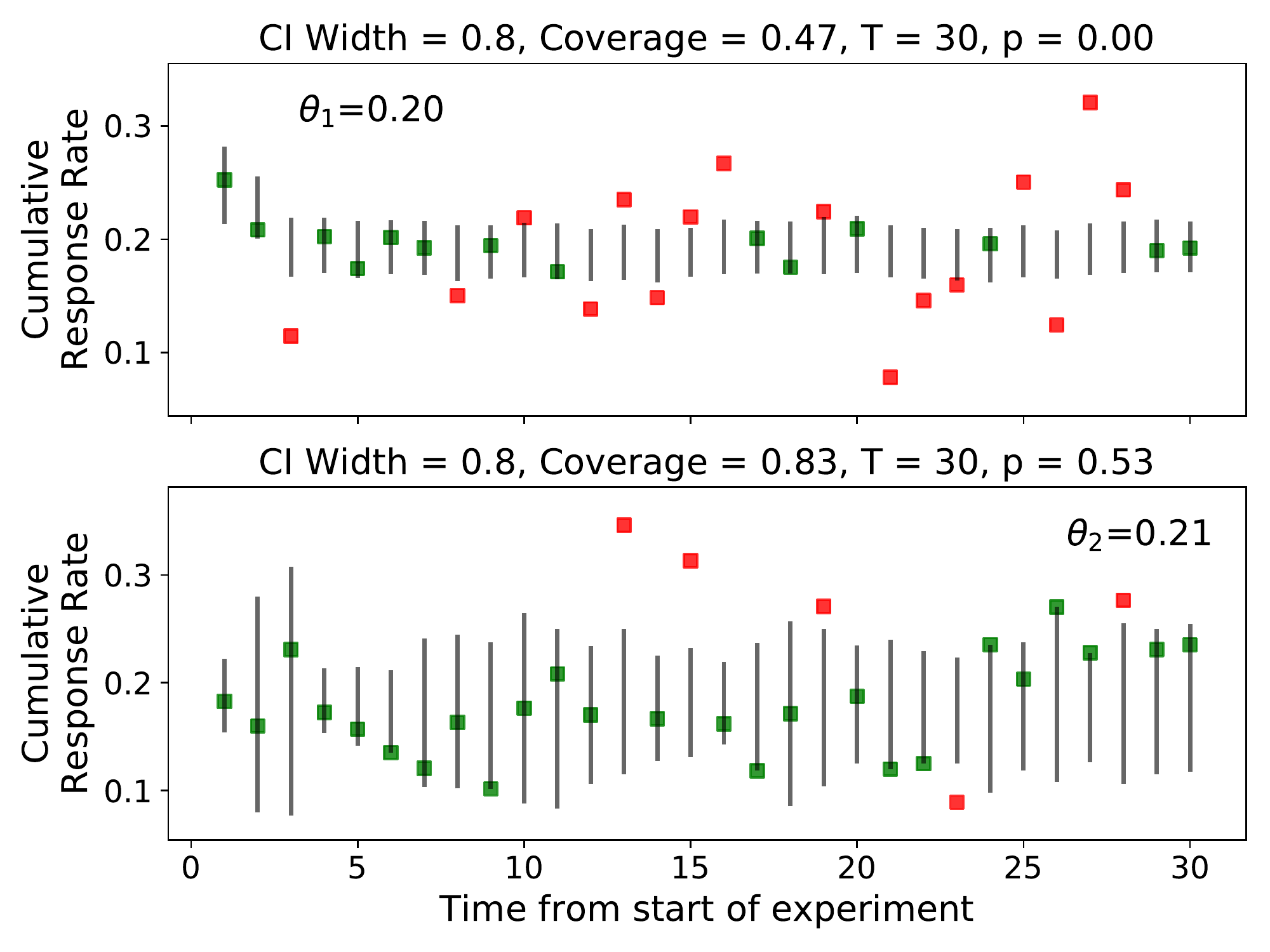}
        \caption{Logistic coverage shows strong evidence of overconfidence.}
        \label{fig:fixed-logistic-coverage}
    \end{subfigure}
    \caption{Logistic (\ref{fig:fixed-logistic-corrected}, \ref{fig:fixed-logistic-bias},\ref{fig:fixed-logistic-coverage}) and BB-GLM (\ref{fig:fixed-bb},\ref{fig:fixed-bb-coverage}) behavior on the fixed but overdispersed dataset~\ref{fig:fixed-data}.}
    \vskip -0.15in
\end{figure*}
\section{Case Studies in Diagnosing and Resolving Common Challenges for Bayesian Bandits}\label{sec:challenges}

We do not seek to claim optimality but rather to clearly demonstrate specific cases where standard models are misspecified and how to diagnose them.
Hence, our three case studies use simulations closely matching real-world UXO data~\cite{van_adelsberg_modeling_2019}.
Different agents stop at different times, and some may fail to ever stop.
So we simulate fixed durations that balance simulation cost with ensuring sample sizes sufficient to support stopping by a reasonable method.
Overdispersion is simulated using a generative process similar to the likelihood for BB-GLM.
For succinctness, we exclude further simulation details but release the data, along with code for generating it~\cite{sweeney_2020_3836504}.

\subsection{Overdispersion due to Unobserved Heterogeneity}\label{subsec:heterogeneity}

Fig.~\ref{fig:fixed-data} shows a simulated dataset with fixed rates $\theta_1 = 0.2, \theta_2=0.21$, and overdispersion induced by generating daily counts from a Beta-Binomial process.
Two common patterns emerged for IBB and Logistic -- both misspecified for the extra-Binomial variance in this dataset.
In some cases (Fig.~\ref{fig:fixed-logistic-corrected}), overdispersion leads to initial overconfident exploitation but $\hat{\theta}_1$ drops enough to restore exploration to the superior UX within the time horizon of the experiment.
In others (Fig.~\ref{fig:fixed-logistic-bias}), exploration is not restored.

With stationary rates and a long enough time horizon, both models will eventually converge on the optimal policy.
If a sub-optimal UX is getting all the traffic, its rate estimate will eventually converge to the true rate.
As this happens, traffic will shift to the UX(s) viewed as inferior and their estimates will converge in turn.
Even so, the desire for rapid iteration through any-time stopping motivates a correction for this overconfidence.
Had we allowed the agent to optionally stop by thresholding the arm probabilities at 95\%, $a_1$ would have been chosen on the first day in both cases.

A common approach for alleviating overconfidence is to force equal allocation and prevent optional stopping for the first $N$ days of an experiment.
We repeated the experiments with Logistic and IBB with forced waiting periods of 7 and 10 days.
A 7-day wait sometimes prevents overconfidence in the Logistic model but never for IBB, indicating the hierarchical shrinkage has the desired effect.
A 10-day wait was always sufficient for both models.

We can either be conservative at a fixed opportunity cost or else be aggressive and increase the risk of sub-optimal stopping.
Alternatively, we can use PPCs to diagnose overdispersion and use a more robust model.
The visually low coverage and the PP $p$-value of 0 for $a_1$ shown in Fig.~\ref{fig:fixed-logistic-coverage} clearly indicate poor calibration.~\footnote{Coverage for $a_2$ appears reasonable only because few samples have been observed for $a_2$, since the policy heavily favors $a_1$.}
Absence of a pattern in which points are above or below the CIs motivates the BB-GLM as a robust yet still stationary extension of the Logistic.

\begin{figure*}[t]
    \centering
    \captionsetup[subfigure]{aboveskip=1pt,belowskip=2pt}
    \begin{subfigure}[t]{0.35\textwidth}
        \vskip 0pt
        \centering
        \includegraphics[width=1\textwidth,height=3.3cm]{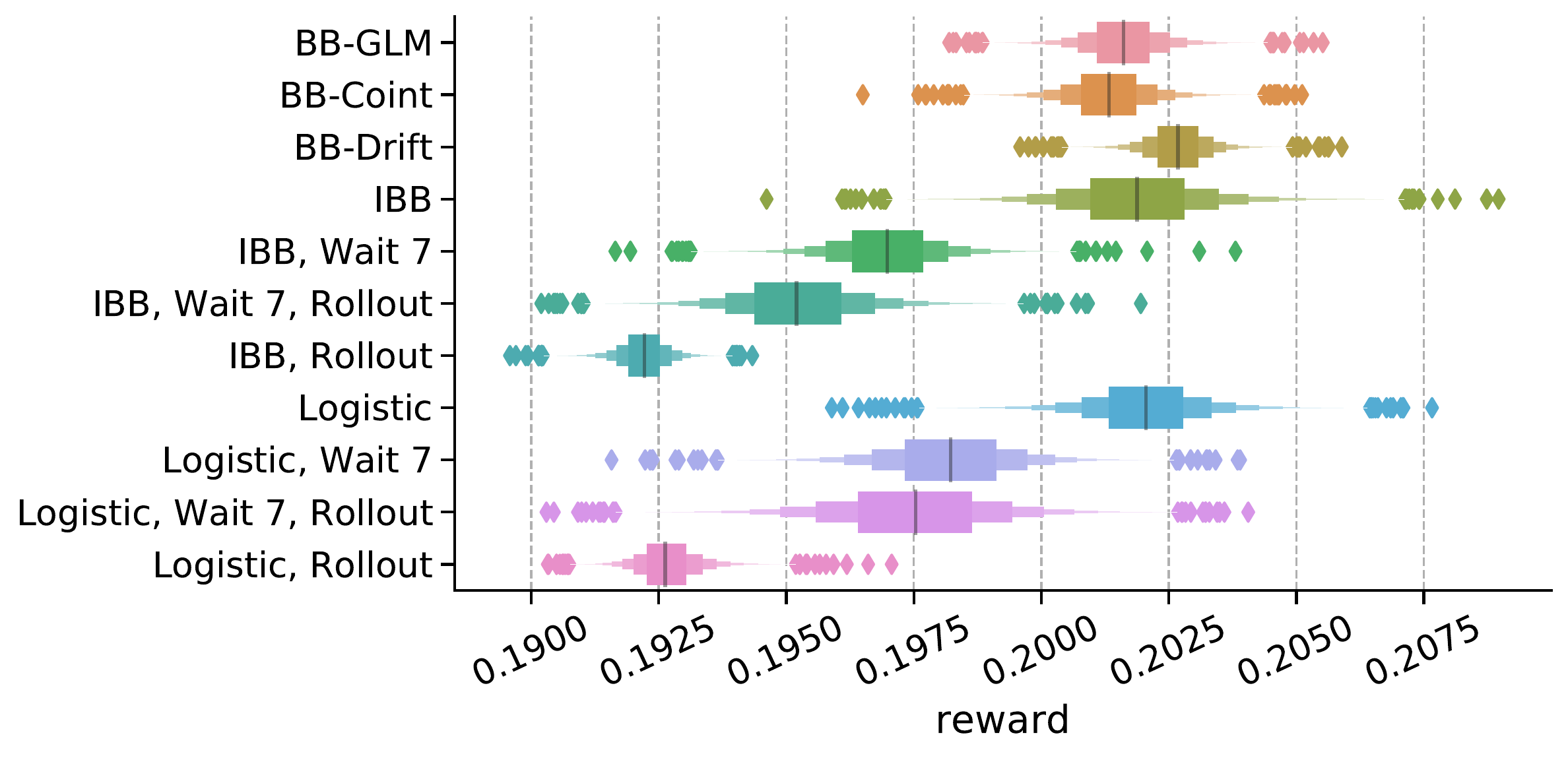}
        \caption{Method comparison on the fixed, overdispersed dataset.
                 BB-GLM, BB-Drift, and BB-Coint with rollout perform the same as without.}
        \label{fig:fixed-comparison}
    \end{subfigure}%
    \hfill
    \begin{subfigure}[t]{0.31\textwidth}
        \vskip 0pt
        \centering
        \includegraphics[width=\textwidth, height=3.3cm]{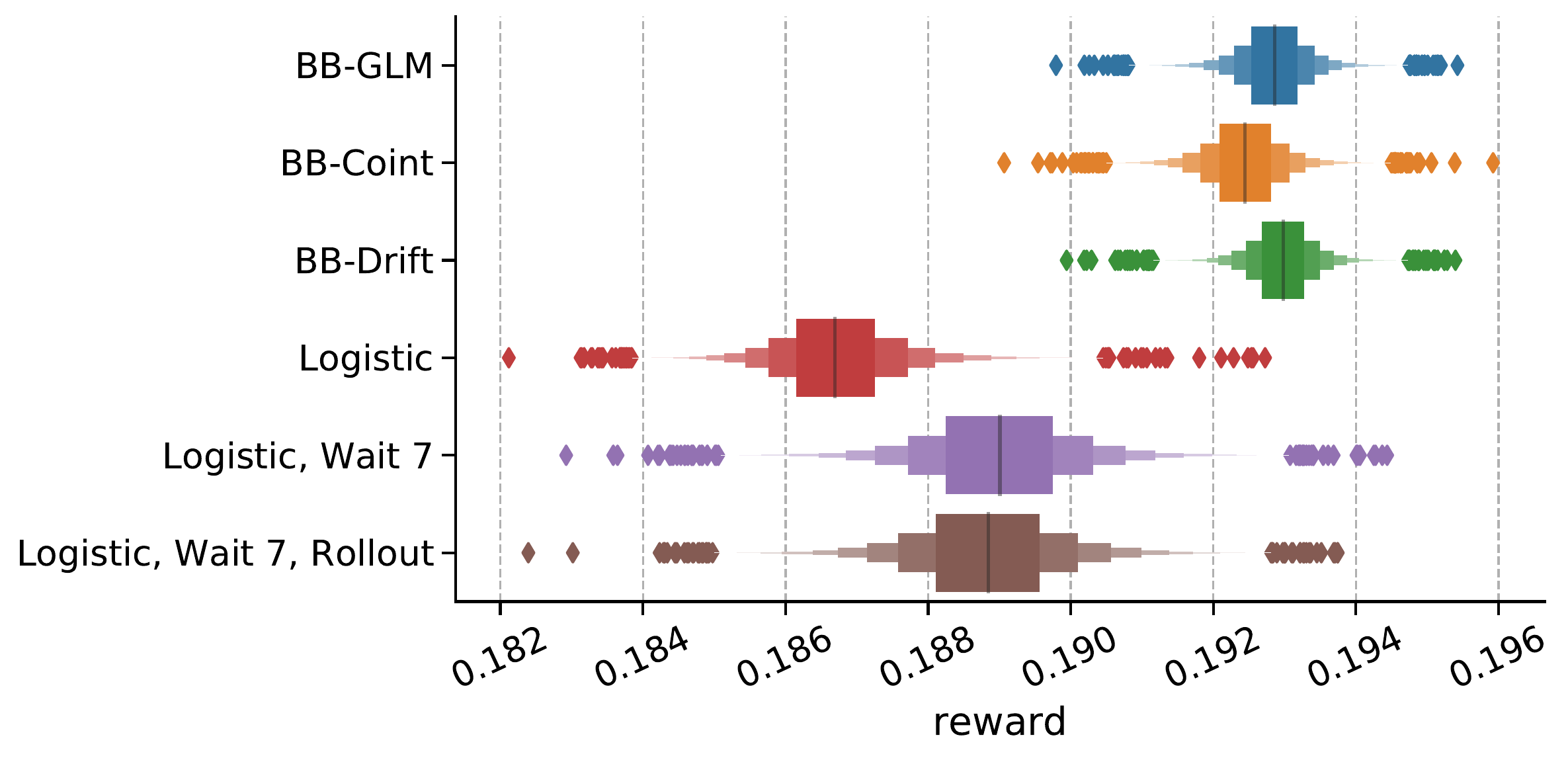}
        \caption{Comparison of BB-GLM, BB-Drift, BB-Coint, and Logistic on the drifting dataset.}
        \label{fig:drift-comparison}
    \end{subfigure}
    \hfill
    \begin{subfigure}[t]{0.31\textwidth}
        \vskip 0pt
        \centering
        \includegraphics[width=1\textwidth,height=2.9cm]{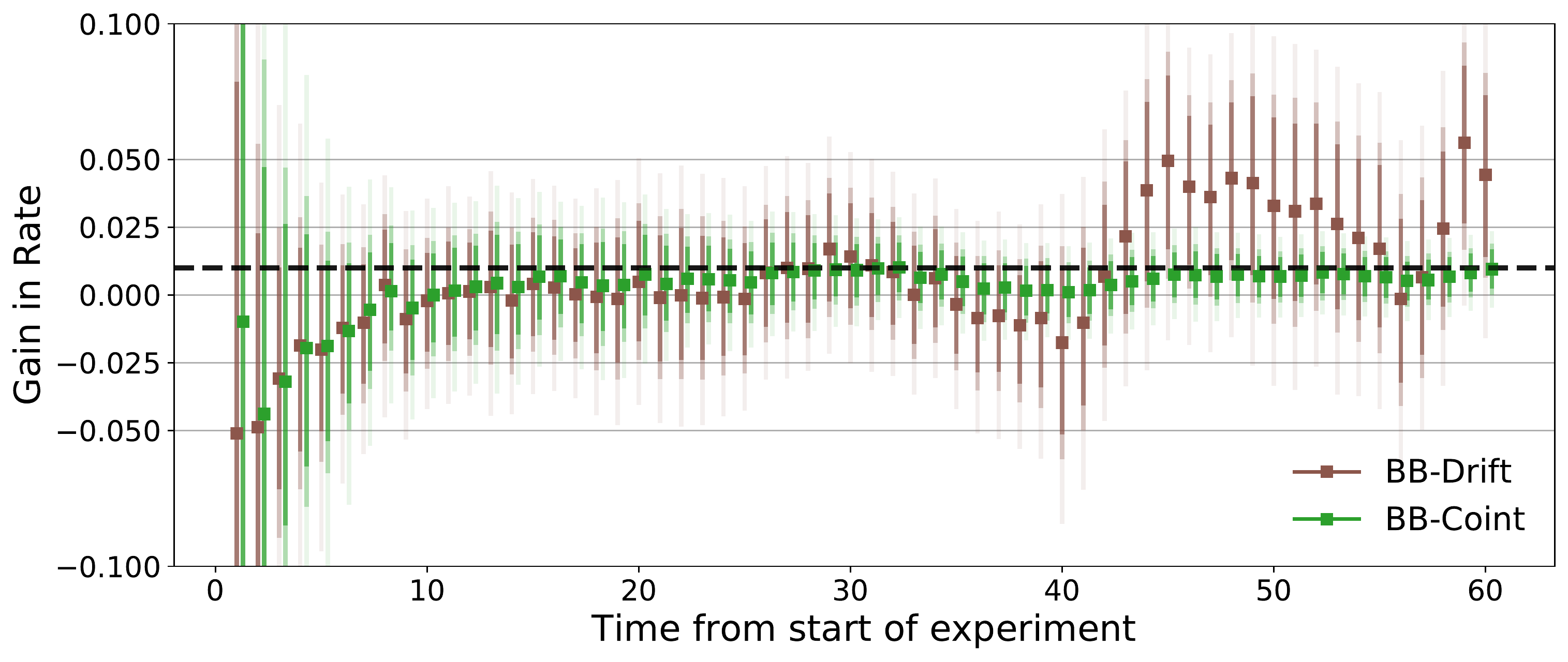}
        \caption{BB-Drift and BB-Coint gain estimates $\theta_2 - \theta_1$ on the drifting dataset.
                 Error bars are 68\%, 80\%, and 95\% CIs.
                 The dotted horizontal line is the true gain of 0.01.}
        \label{fig:drift-gains}
    \end{subfigure}
    \caption{Key comparisons of the various methods on the first two case studies.}
    \vskip -0.05in
\end{figure*}

As seen in Fig.~\ref{fig:fixed-bb}, BB-GLM is robust to the noisy data, swapping the majority of visitors to the better UX as soon as its superiority is evident.
The early conservatism seen in Fig.~\ref{fig:fixed-bb-coverage} allows stopping without concern for overconfidence, while preserving calibration.
Fig.~\ref{fig:fixed-comparison} shows the rewards from DR-ns for BB-GLM, IBB, and Logistic, with and without stopping and heuristics.
Logistic performance is amongst the best on average but is more volatile.
With optional stopping (rollout), its rewards are the worst by far, and if waiting 7 days, it still performs worse than BB-GLM.
%

\subsection{Overdispersion due to Drifting Rates}\label{subsec:drifting}

Section~\ref{subsec:heterogeneity} clearly illustrates the overconfidence of standard models in the presence of unobserved heterogeneity.
Overdispersion also arises from time-dependent response rates.
In contrast to the eventual convergence to an optimal policy in the case of stationary rates, this case study demonstrates the models seen so far can all accumulate persistent bias when rates are drifting.
With Logistic and IBB, this manifests as poor/premature rollout decisions, even after a forced waiting period, and with BB-GLM, as a lasting inability to build confidence in a stable winner.
We now illustrate these limitations and explore whether BB-Drift and BB-Coint can overcome them.

\subsubsection{Bias due to insufficient randomization across time}

\begin{figure*}[t]
    \centering
    \captionsetup[subfigure]{aboveskip=2pt,belowskip=2pt}
    \begin{subfigure}[t]{0.32\textwidth}
        \vskip 0pt
        \centering
        \includegraphics[width=1.05\textwidth,height=2.6cm]{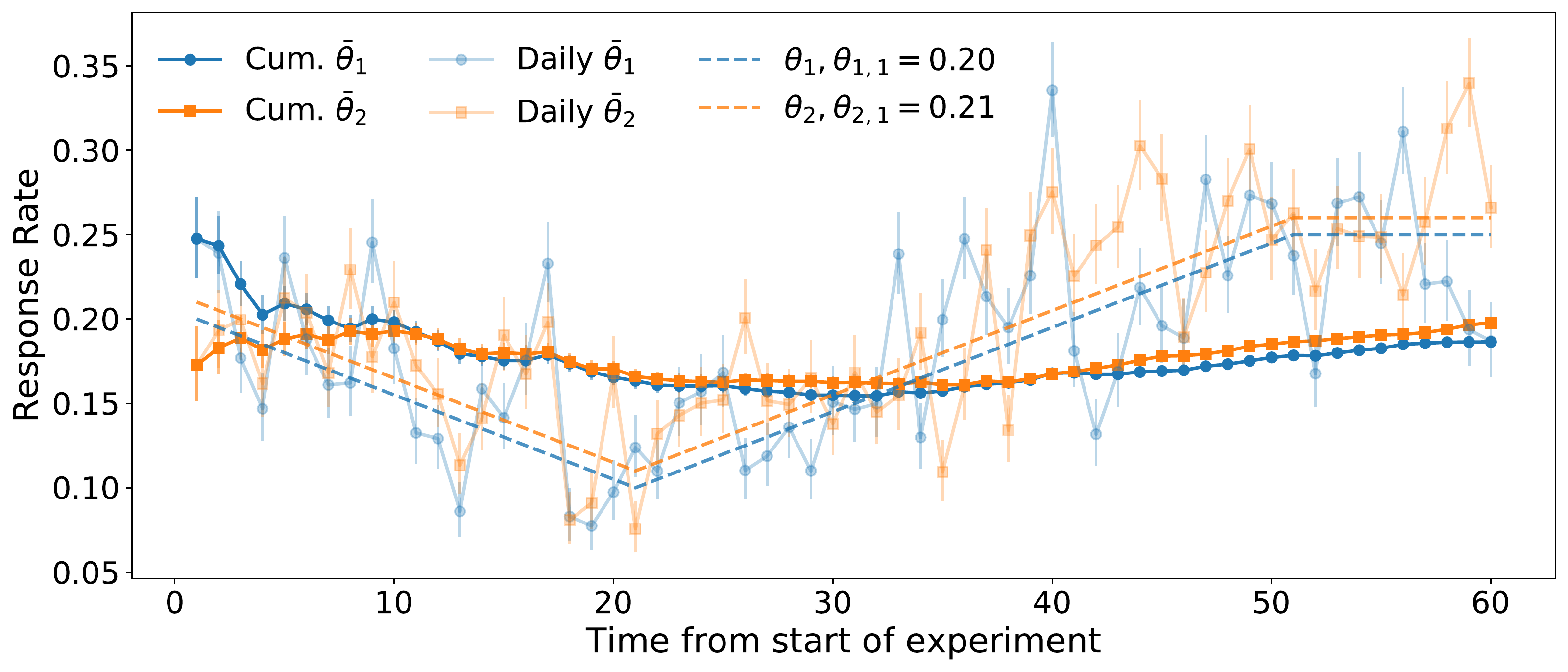}
        \caption{The ``drifting," overdispersed dataset, with a 0.01 gain and 500 samples/day per UX. The rates decrease for 20 days then increase for 30, by 0.005/day, then remain constant.}
        \label{fig:drift-rates}
    \end{subfigure}%
    \hfill
    \begin{subfigure}[t]{0.32\textwidth}
        \vskip 0pt
        \centering
        \includegraphics[width=1.05\textwidth,height=3.3cm]{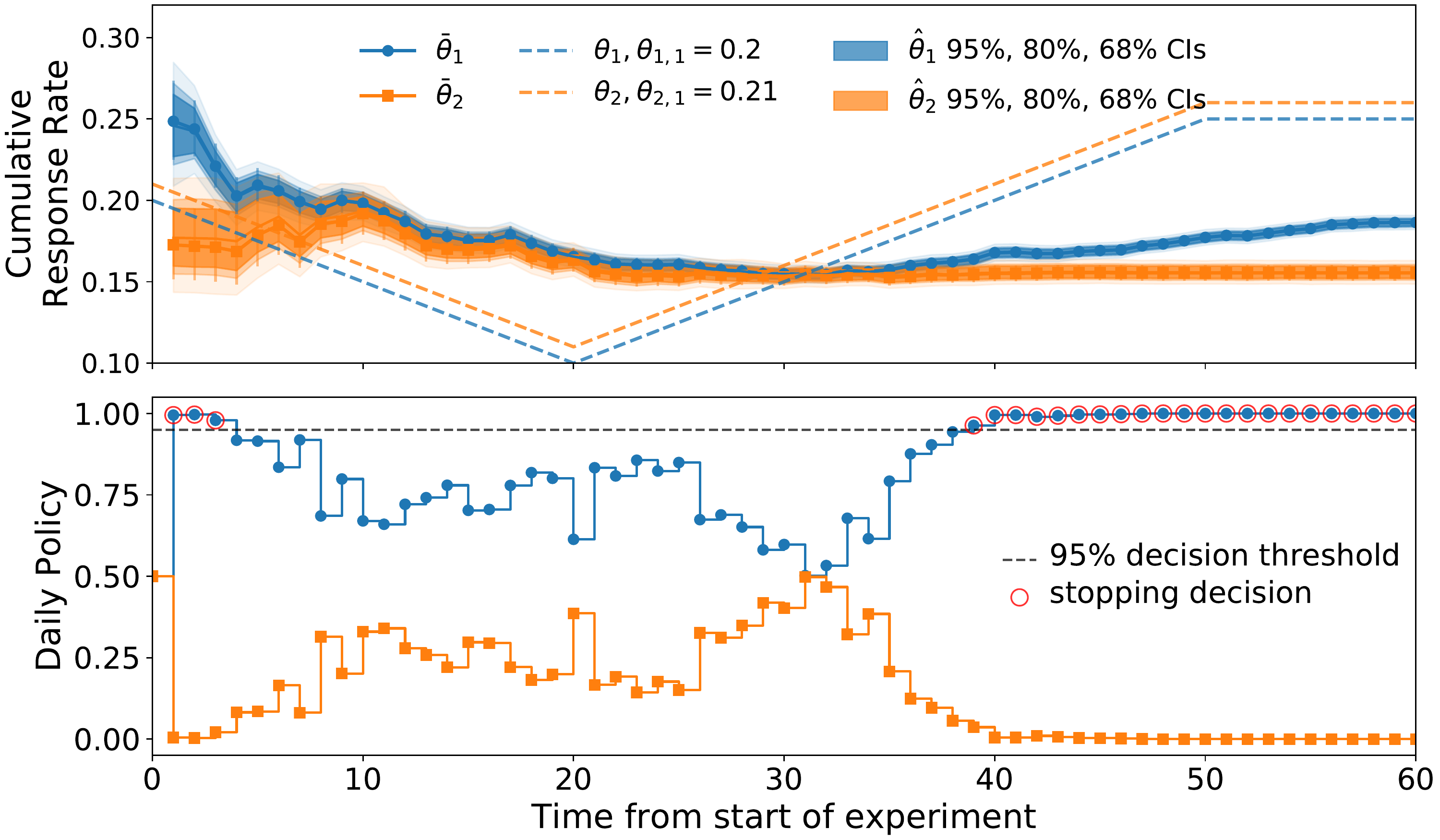}
        \caption{Logistic incurs lasting bias towards $a_1$ from favoring it during the initial uptime.}
        \label{fig:drift-logistic}
    \end{subfigure}%
    \hfill
    \begin{subfigure}[t]{0.32\textwidth}
        \vskip 0pt
        \centering
        \includegraphics[width=1.05\textwidth,height=3.3cm]{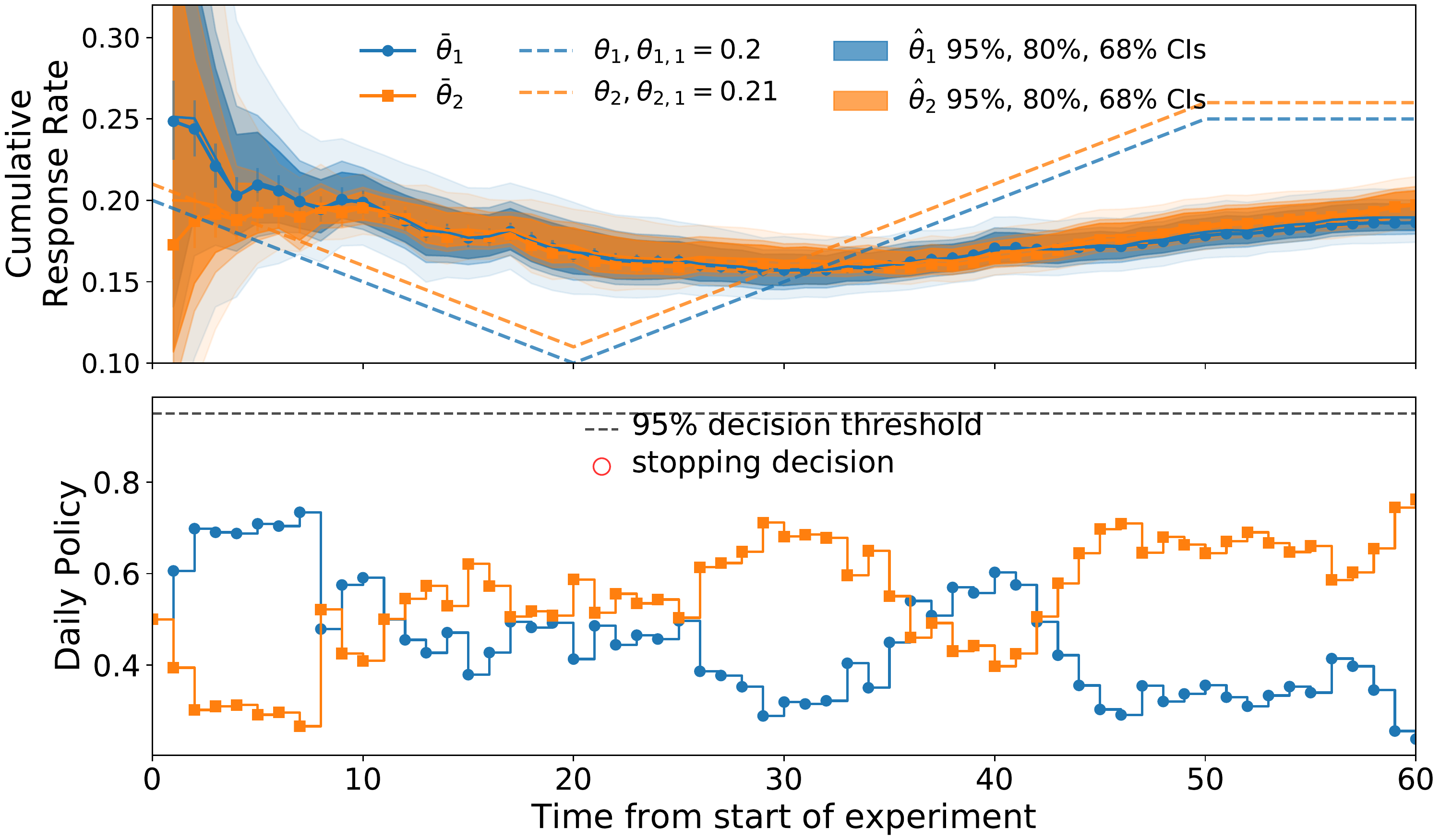}
        \caption{BB-GLM is more robust than Logistic but is slow to learn since it sees drift as noise.}
        \label{fig:drift-bb}
    \end{subfigure}%
    \hfill
    \begin{subfigure}[t]{0.32\textwidth}
        \vskip 0pt
        \centering
        \includegraphics[width=\textwidth,height=3.3cm]{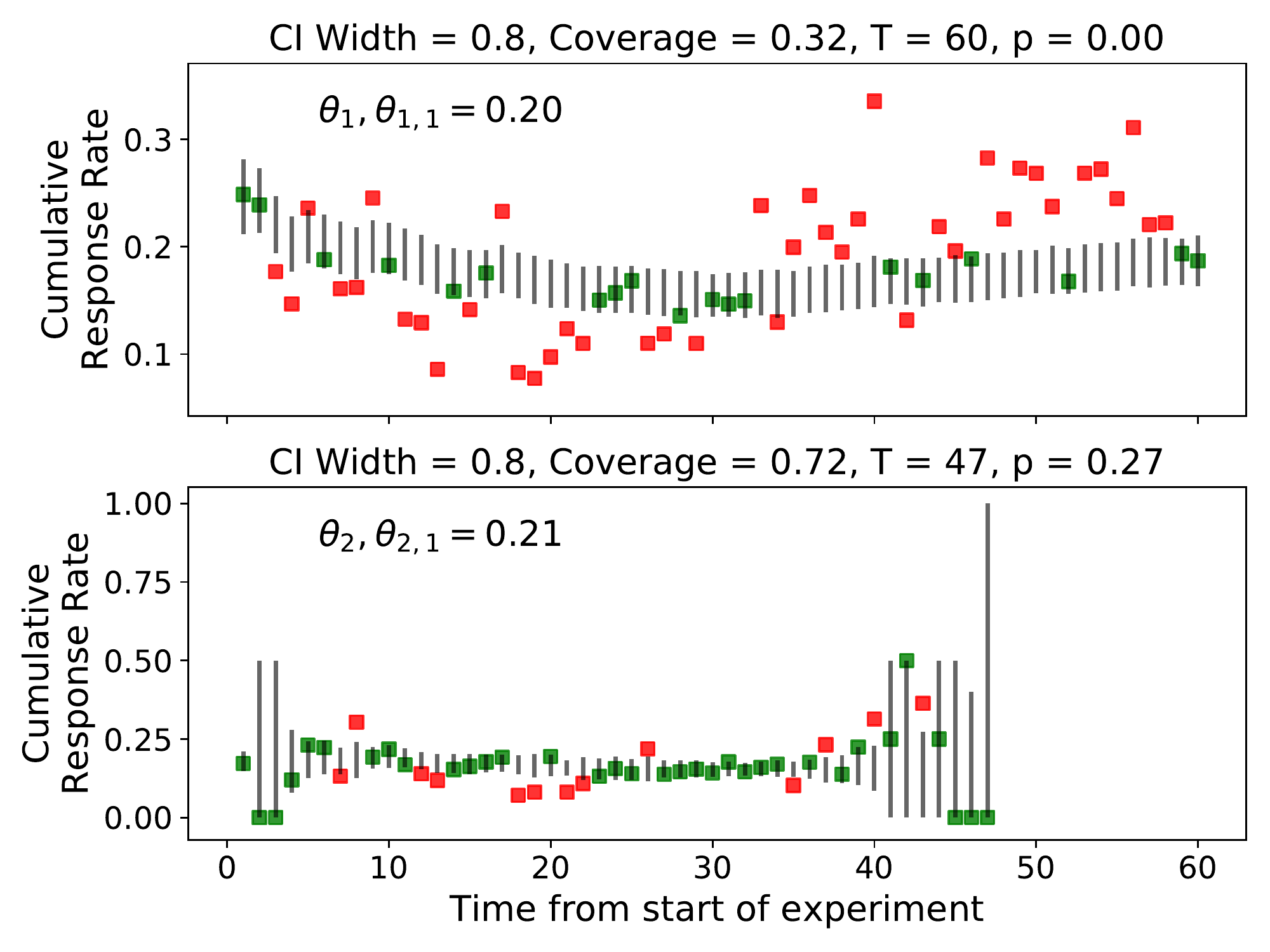}
        \caption{Logistic coverage shows clear evidence of overconfidence and time-dependence.}
        \label{fig:drift-logistic-coverage}
    \end{subfigure}%
    \hfill
    \begin{subfigure}[t]{0.32\textwidth}
        \vskip 0pt
        \centering
        \includegraphics[width=1.05\textwidth,height=3.3cm]{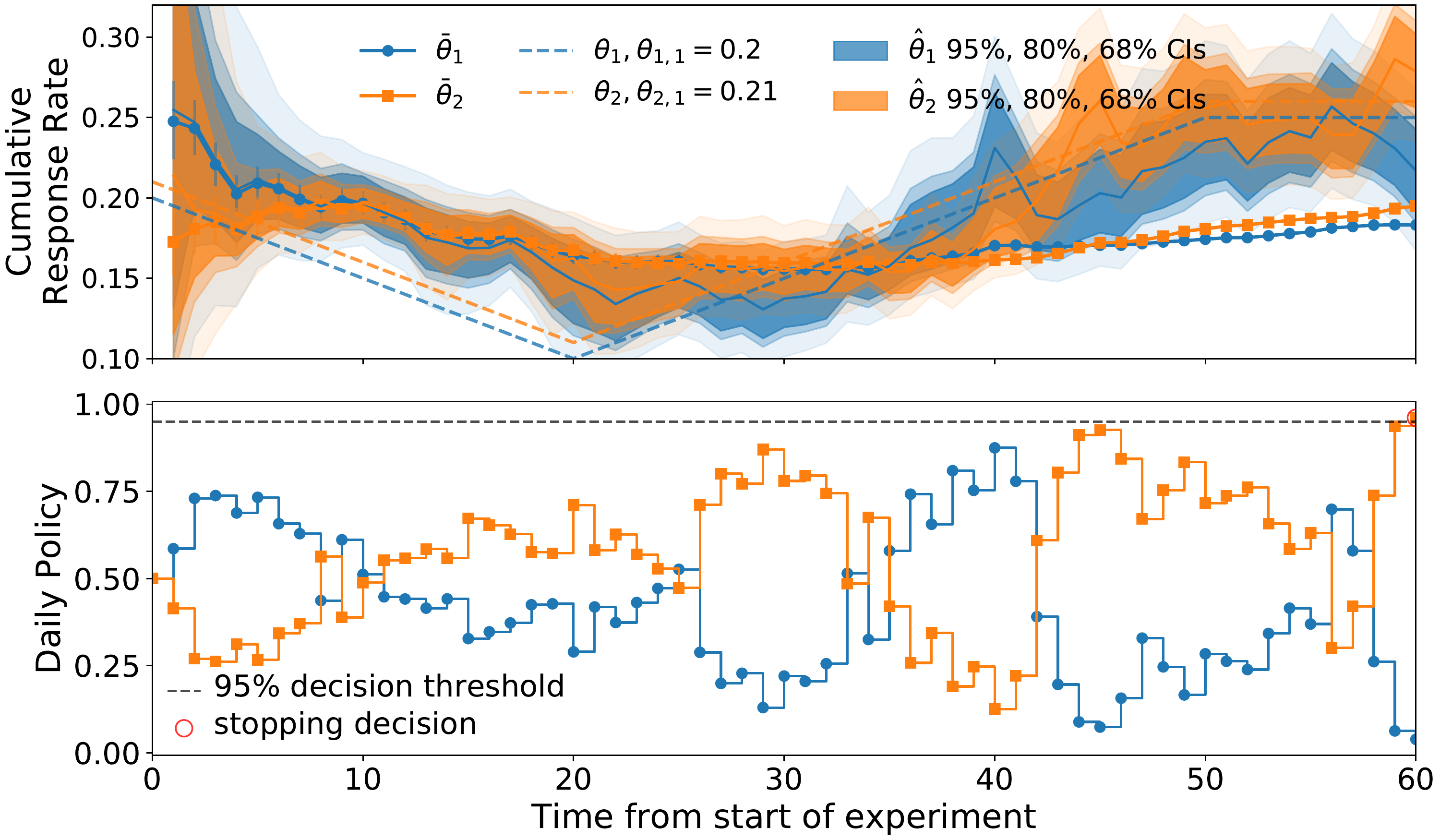}
        \caption{BB-Drift discounts older data to avoid bias, but its volatility even late in the experiment is undesirable.}
        \label{fig:drift-bb-drift}
    \end{subfigure}%
    \hfill
    \begin{subfigure}[t]{0.32\textwidth}
        \vskip 0pt
        \centering
        \includegraphics[width=1.05\textwidth,height=3.3cm]{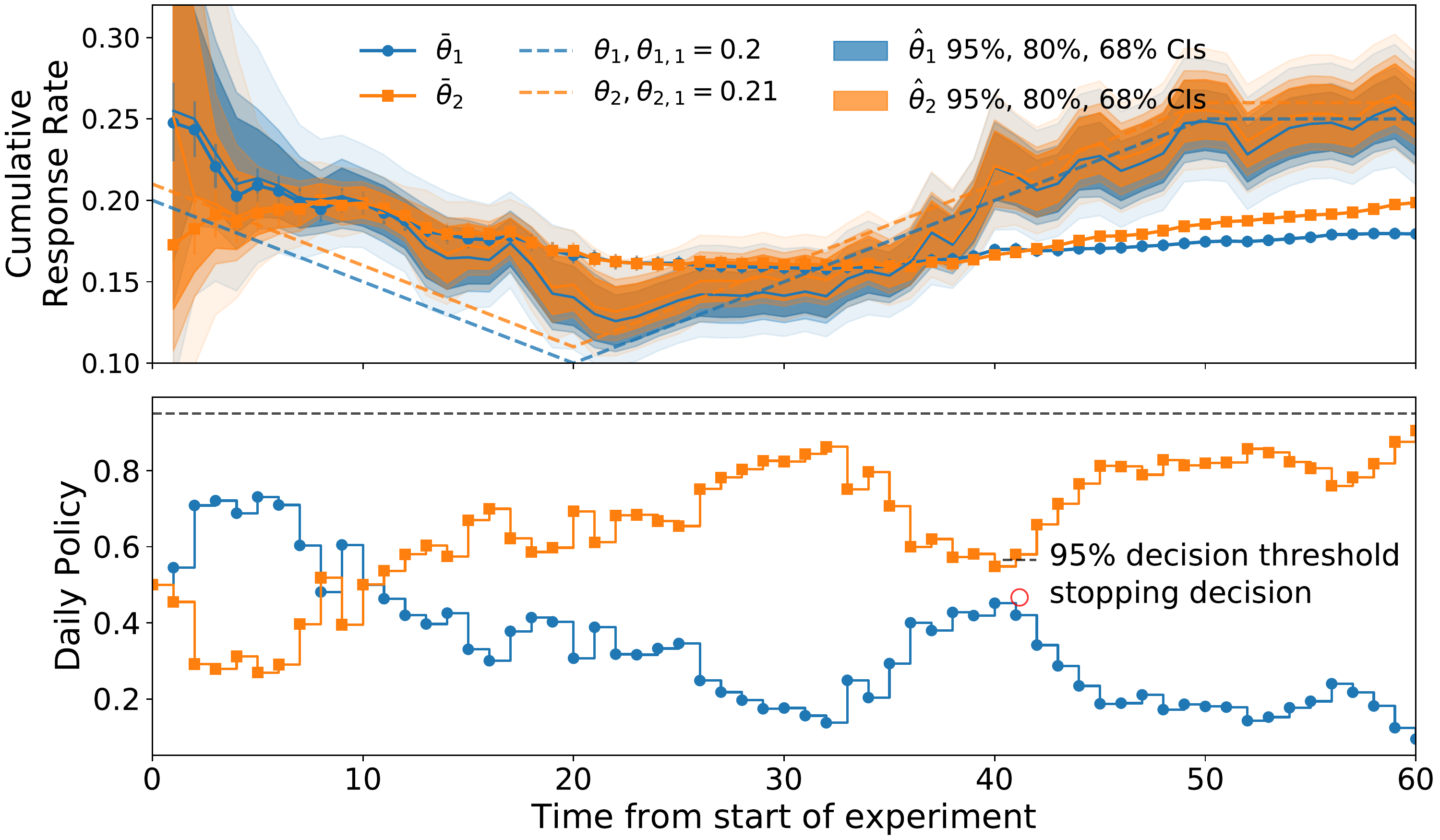}
        \caption{BB-Coint maintains a much more stable policy because of its ability to capture stable gains between UXs.}
        \label{fig:drift-bb-coint}
    \end{subfigure}
    \caption{Logistic, BB-GLM, BB-Drift, and BB-Coint on the overdispersed, drifting dataset~\ref{fig:drift-rates}.}
    \vskip -0.15in
\end{figure*}

Fig.~\ref{fig:drift-rates} shows a dataset with two UXs whose true rates begin at $\theta_{1,1}=0.20, \theta_{2,1}=0.21$, are decreased by 0.005/day for 20 days, increased by 0.005/day for 30, then held stable for 10.
This simulates drift where the true gain remains positive throughout, which is necessary for a stable winner to exist.
Representative repetitions of Logistic and BB-GLM are shown in Fig.~\ref{fig:drift-logistic} and Fig.~\ref{fig:drift-bb};
we exclude IBB since its limitations are simply a more severe version of the Logistic.

As in the previous case, Logistic overconfidently allocates most visitors to $a_1$, which appears best during the first few days.
By day four, the estimate drops enough for the agent to slowly start exploring $a_2$, but by this time, the rates have already drifted downwards.
So while the daily rates are lower for both UXs, the policy has already produced a sample biased towards $a_1$.
Essentially, the Logistic agent got to see $a_1$ during the good times but only starts to see $a_2$ during the down times, and it has no idea that conditions have changed.
As seen in Fig.~\ref{fig:drift-bb}, BB-GLM is less overconfident but still initially favors $a_1$, incurring some bias that makes it slow to build confidence in a winner later.

We can diagnose these issues in several ways.
In Fig.~\ref{fig:drift-logistic-coverage}, Logistic overdispersion is evident, but we can also see more points fall under the CI early on while more are over later, indicating nonstationarity.
We can also conduct an \textbf{autocorrelation} (AC) PPC for each UX.
AC is the lag-1 correlation ($\in [-1, 1]$) of a series with itself.~\cite{nist_sematech_2013}
Values further from 0 indicate drift is more likely.
The AC for both variants is above 0.5, while BB-GLM believes both close to 0, resulting in PP $p$-values of 0.

We again consider heuristic and model-based solutions.
BB-Drift AC PP $p$-values of 0.76 and 0.6 for the two variants indicate good fit, and Fig.~\ref{fig:drift-bb-drift} shows BB-Drift remains skeptical early on, as in BB-GLM, but much more confidently exploits $a_2$ further in.
However, the volatility of the policy based on BB-Drift near the end of the experiment is another issue, which we now address.

\subsubsection{Drift model's inability to identify stable gains}\label{subsubsec:cointegration}

As we saw in the previous section, modeling both drift and extra-Binomial dispersion can prevent overconfidence and bias without tuned heuristics.
However, BB-Drift has a shortcoming.
We expect more confidence with more data, but Fig.~\ref{fig:drift-gains} shows the uncertainty of the BB-Drift gain estimates $\theta_2 - \theta_1$ remain large and volatile, actually going negative just days before the agent declares a winner.
As seen in Fig.~\ref{fig:drift-bb-coint} and Fig.~\ref{fig:drift-gains}, BB-Coint maintains a more stable policy than BB-Drift while also maintaining more confidence than BB-GLM.

BB-GLM, BB-Drift, and BB-Coint all obtain similar reward rates on this dataset, shown in Fig.~\ref{fig:drift-comparison}.
Logistic is worst even with a forced waiting period.
The similarity in rewards of the first three models hides all the nuance of the qualitative analysis above.
With more and longer datasets, we might see more differentation, but this example indicates \textit{NPE should not be the exclusive method for agent comparison}.

\begin{figure*}[t]
    \centering
    \captionsetup[subfigure]{aboveskip=1pt,belowskip=2pt}
    \begin{subfigure}[t]{0.36\textwidth}
        \vskip 0pt
        \centering
        \includegraphics[width=\textwidth,height=2.95cm]{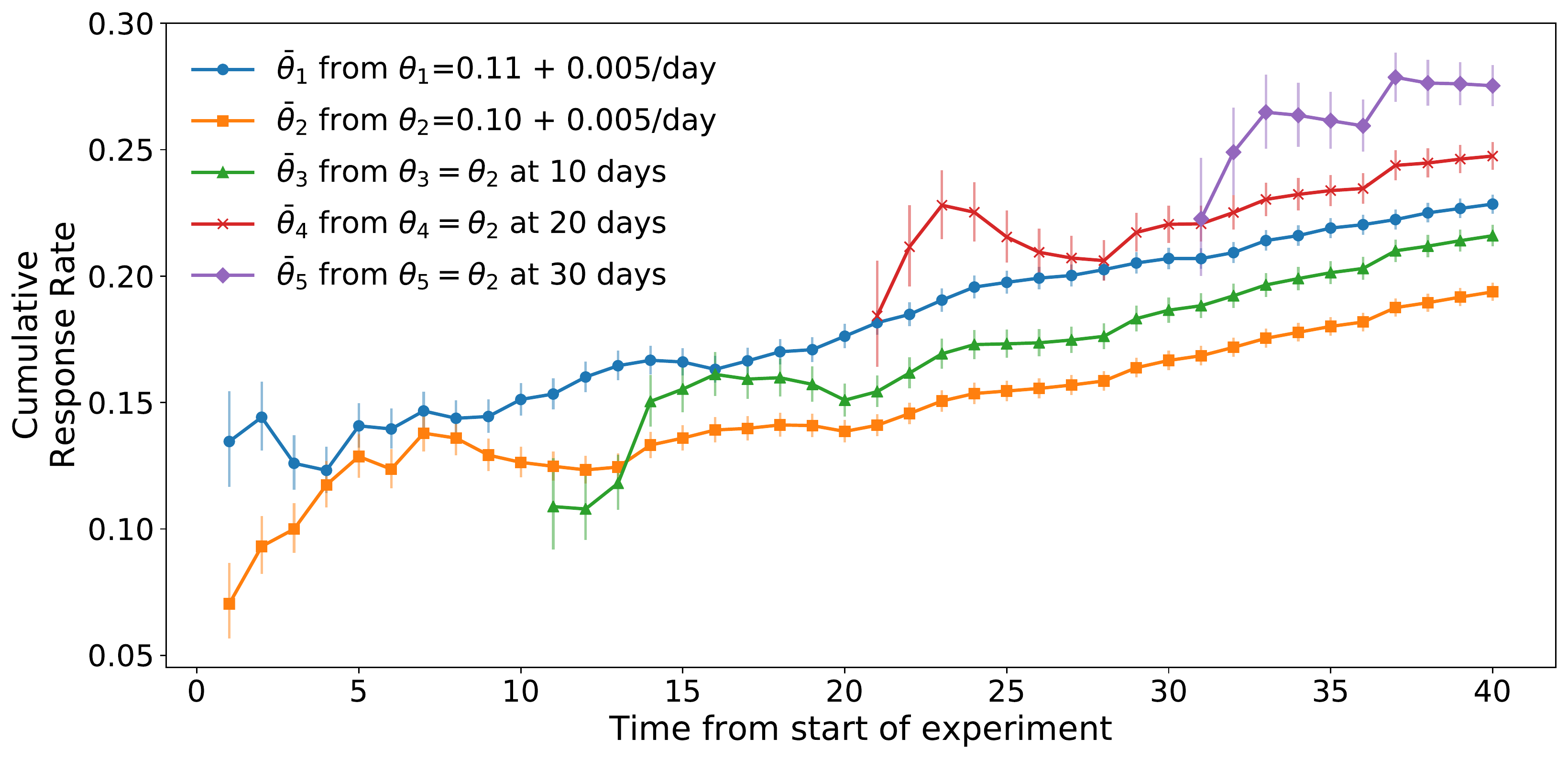}
        \caption{Dataset with ``drifting" effects, overdispersion, a 0.01 gain, and arrival rates of 500 visitors per UX per day.
                 $a_2$ duplicates are introduced at 10, 20, and 30 days.}
        \label{fig:arm-add-rates}
    \end{subfigure}%
    \hfill
    \begin{subfigure}[t]{0.3\textwidth}
        \vskip 0pt
        \centering
        \includegraphics[width=1.05\textwidth,height=3.3cm]{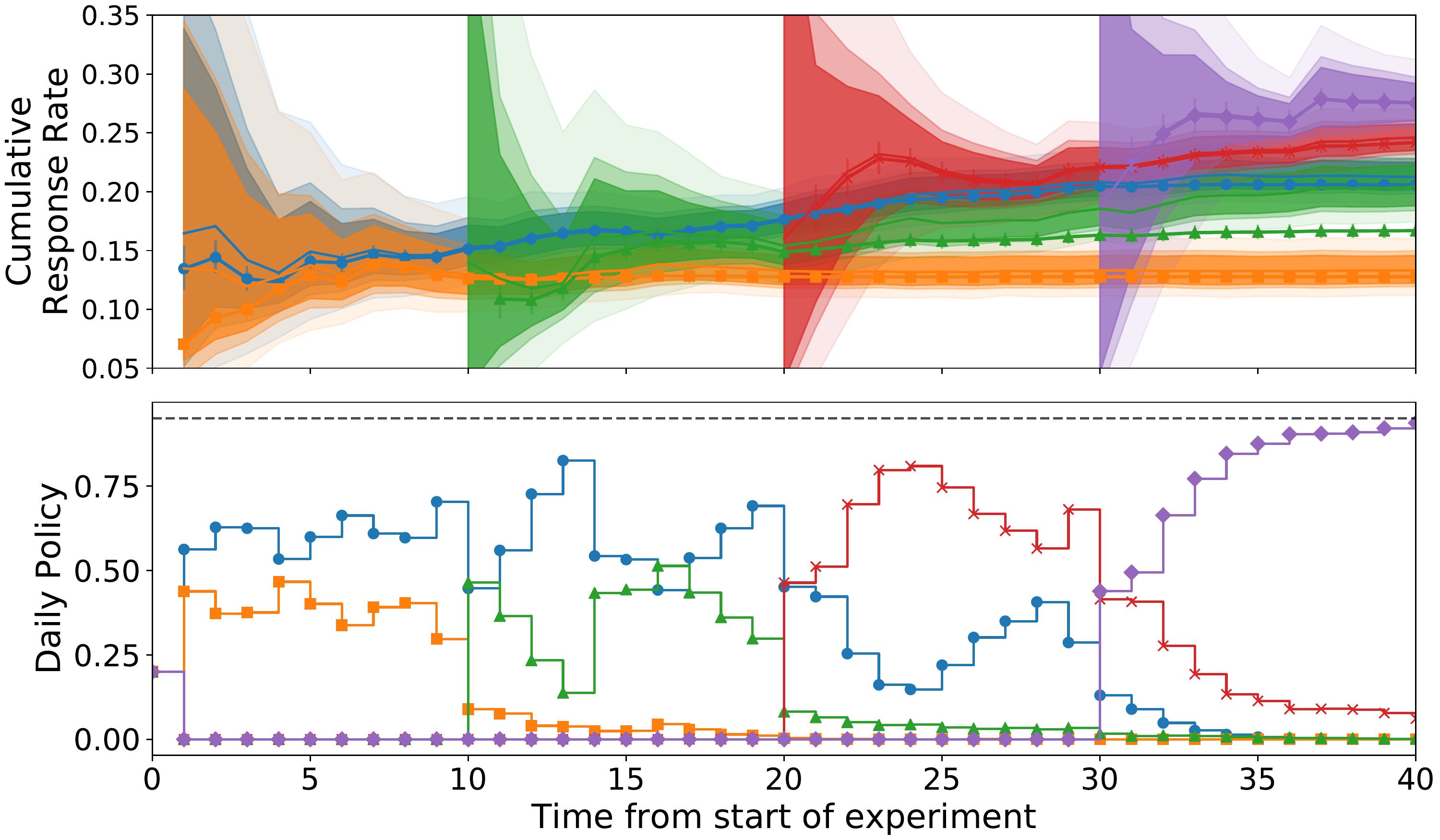}
        \caption{BB-GLM is biased towards the added arms.}
        \label{fig:arm-add-bb}
    \end{subfigure}
    \hfill
    \begin{subfigure}[t]{0.3\textwidth}
        \vskip 0pt
        \includegraphics[width=1.05\textwidth,height=3.3cm]{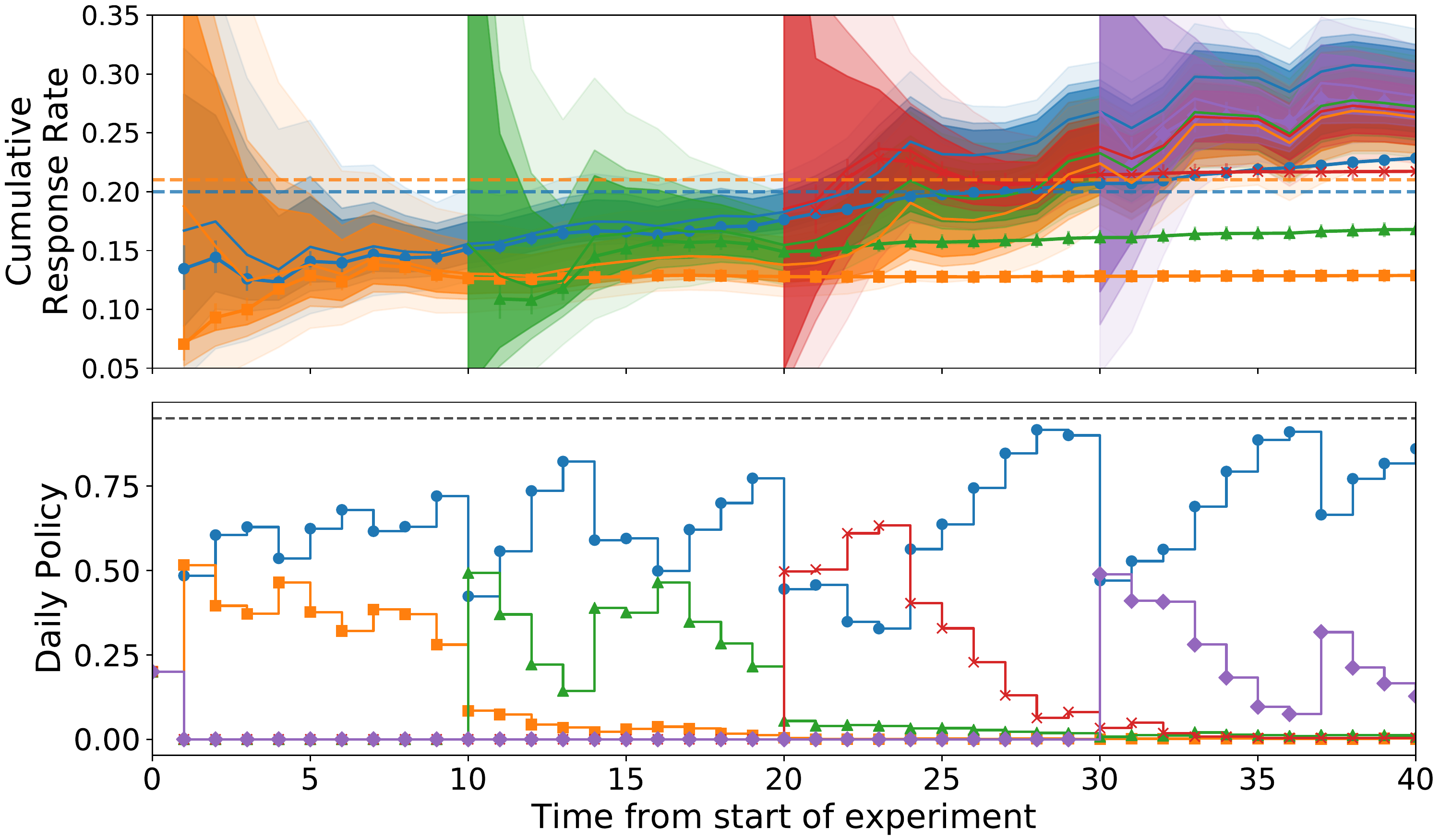}
        \caption{BB-Coint quickly discards each new UX as inferior to the control.}
        \label{fig:arm-add-bb-coint}
    \end{subfigure}
    \caption{BB-GLM and BB-Coint on the arm addition dataset~\ref{fig:arm-add-rates}.}
    \vskip -0.2in
\end{figure*}

\subsection{Bias When Adding Arms When Rates Are Drifting}\label{subsec:warm-start}

We now consider the case where new UXs are added on-demand during an experiment.
When a winner is chosen, a new set of UXs is created and introduced into the test alongside the old ones.
This is commonly handled by restarting the experiment, but this cold-start approach can drastically limit the number of variants that can be tested.
So it is desirable to continue the experiment and view preceding observations for the new UXs as missing.
With this approach, extreme bias can arise when rates are drifting but the value model assumes stationarity.

Fig.~\ref{fig:arm-add-rates} shows a simulated dataset that starts with two UXs with rates 0.10 and 0.11 that increase by 0.005/day.
Think of $a_1$ as the previous "winner," which we are now iterating on by introducing $a_2$.
We introduce three duplicates of $a_2$ at 10, 20, and 30 days.
All models view these as new, but their data are identical to $a_2$, minus observations before their introduction.
Note how the empirical cumulative rates of $a_3$, $a_4$, and $a_5$ are successively higher due to the upward drift.

We compare BB-GLM to BB-Coint, since previous cases indicate they are superior versions of Logistic and BB-Drift.
As seen in Fig.~\ref{fig:arm-add-bb}, BB-GLM is increasingly biased towards each new addition, while BB-Coint (Fig.~\ref{fig:arm-add-bb-coint}) considers but quickly rules out each in turn.
BB-Drift (not pictured) performs better than BB-GLM but struggles to decide which UX is best near the end; the uncorrelated drift dynamics erode confidence in estimates for previously ruled out UXs.
So properly handling warm-start when adding arms appears to require modeling both time-awareness and cointegration.

Also note that this example shows how models assuming stationarity will be biased towards new UXs when rates have been drifting upwards, but the opposite can also occur (bias against new additions when rates are drifting down).
\section{Discussion and Recommendations}\label{sec:discussion}

\subsection{Summary of Case Studies}

Through several case studies, we demonstrated that the commonly used Logistic and conjugate Beta-Binomial models suffer from severe overconfidence in the presence of overdispersion.
In the case of overdispersion due to unobserved heterogeneity, the models are prone to initial overconfidence but will eventually correct their beliefs.
However, when optional stopping is used, a rollout decision may be made before this correction occurs.
Application of forced exploration rules and a decision waiting period can make this less likely, but such heuristics simply trade one form of opportunity cost for another, as they must be carefully tuned.
An alternative approach is to use a robust model such as our BB-GLM, which allows the agent to learn the right exploration factor based on rate volatility between time units.

In the case of overdispersion due to drifting rates, neither heuristic rules nor robust models can fully prevent bias due to incomplete randomization across time.
Introducing drift dynamics to the BB-GLM provides model-based discounting that prevents this bias, but the inability of this model to capture cointegration prevents it from confidently identifying stable winners.
Extending the drift model to capture cointegration corrects these issues, and this model (BB-Coint) is also able to seamlessly handle the addition of new UXs while warm-starting from previously observed data, whereas the Logistic and BB-GLM can easily be biased towards/against new UXs when rates are drifting up/down.

\subsection{Is There One Model to Rule Them All?}\label{subsec:model-rec}

Given the limitations of Logistic and BB-GLM, it's natural to wonder if we should always prefer BB-Coint.
The concern for bias in the restless bandit with drifting rates clearly motivates its use, but what if there is no drift?
Figure~\ref{fig:fixed-comparison} shows a comparison of Logistic, BB-GLM, and BB-Coint on the first case, which has no drift.
We already saw that Logistic is outperformed by BB-GLM when optional stopping is at play, as extra exploration must be forced to mitigate its overconfidence.
Since BB-Coint and BB-GLM perform nearly identically, it appears BB-Coint can be a good option even when there is no drift.

Of course, our approach here has just focused on a few case studies to clearly illustrate how specific value model misspecifications can impact bandit and optional stopping behavior; we do not seek to make broad claims of optimality.
Even so, we recognize that a practitioner must make modeling decisions with incomplete information, and that, as Box famously said "...all models are wrong, but some are useful"~\cite{Box_1976}.
At a minimum, we strongly advise model checking when employing one of the simpler models such as IBB and Logistic.
If this checking indicates you are facing the issues we have demonstrated here, consider one of the following three strategies for addressing them.

\subsection{Strategies for Mitigating Model Misspecification Risks}

\begin{enumerate}
    \item \textbf{More covariates}. Include additional covariates in the Logistic model.
    \item \textbf{Start general}. Employ the most general model and report summaries of parameters which indicate what data characteristics it is uncovering (overdispersion, nonstationarity, etc.).
    \item \textbf{Test-to-complicate}. Employ a suite of progressively more complex models and carefully selected model checks whose evaluation collectively prescribe the simplest adequate model for a given environment.
\end{enumerate}

\subsubsection{More Covariates}

The first approach considered should always be including additional covariates in a simple model -- Logistic Regression in this case -- to control for confounding.
This can make both overdispersion and nonstationarity disappear, while also providing deeper insight into what factors drive change.
Even if data is not currently available, seeking out additional data that seem likely to explain observed variance and time-dependence is often a better use of time and effort than developing or validating a more complex model.
In this paper, we started by assuming this was not an option.
Only once it has been ruled out should the next two options be considered.

\subsubsection{Start General}

The Bayesian paradigm provides a natural mechanism for incorporating uncertainty about model selection, usually termed \textit{continuous model expansion}~\cite{bda3}.
First, embed all models you believe reasonably plausible for your data into one super-model.
Then, select priors which encode your beliefs about how plausible you believe each is.
Finally, let the logic of Bayesian inference take over, incorporating evidence from the data as it arrives.
In this case, selecting the BB-Coint model for every experiment can be seen as an instance of this strategy.

At first glance, this ``robust" modeling approach seems ideal;
it reduces the model checking burden by safeguarding against many possible types of model misspecification in one go;
however, our lunch is never really free~\cite{Wolpert_Macready_1997}.
The most general model will also be the most complex and so the most computationally expensive.
It also requires specification of informative priors about many aspects of the data to avoid inferential issues like identifiability~\cite{koop_bayesian_2006}.
While this is a useful exercise, it may be challenging if there is little data available, and it will be time-consuming up-front.

If deciding to go this route, it is often beneficial to consider what parameter summaries to output to improve understanding of your data characteristics over time.
Let questions like these drive these decisions:
\begin{itemize}
    \item Is the posterior dispersion parameter large, indicating significant overdispersion?
    \item How much is the data drifting between time steps?
\end{itemize}

\subsubsection{Test-to-complicate}

Instead of embedding all plausible models in one super-model, develop a tree of progressively more complex models, each addressing limitations of one or more of the others, and then employ model checks for each to determine the simplest option which is good enough for your application.
Algorithm~\ref{alg:test-to-complicate} is what this might look like for UXO.
One possible output of this algorithm is that there is no model suitable for the given dataset, in which case one option might be to fall back to a complete randomization policy.

\begin{algorithm}
    \caption{Test-to-complicate for UXO}
    \label{alg:test-to-complicate}
    \begin{algorithmic}[1] 
        \For{\text{model} $M \in$ \{Logistic, BB-GLM, BB-Coint\}}
            \State Fit $M$
            \State Run coverage check per test cell to compute $p$-values
            \If{$p > 0.1$ for all cells}
                \Return fitted $M$
            \EndIf
        \EndFor
        \State \Return ``No suitable model" message
    \end{algorithmic}
\end{algorithm}

This approach may be less computationally expensive than \textit{Start General} if a simple model is sufficient but will be more expensive if the most complex model is required, as it requires fitting the simpler models first.
If that is not of concern, the main requirement for this approach is defining model checks for ruling out each model.
We are proponents of PPCs but recognize the challenge of defining posterior predictive $p$-values that can be used in an automated fashion.
If the application permits an offline analysis, it is often preferable to combine such checks with a graphical component, as we have done here.
If not, e.g. the model is running in an automated pipeline, or being used for many experiments, more careful consideration of $p$-value calibration may be warranted~\cite{Gneiting_Balabdaoui_Raftery_2007, Hjort_Dahl_Steinbakk_2006}.
\section{Conclusions and Future Work}\label{sec:conclusion}

In this empirical study, we showed several ways in which standard value models are misspecified on many real-world data, leading to sub-optimal rewards.
We provided a novel formulation of UXO as a restless, sleeping bandit with unobserved confounders plus optional stopping -- challenges common in other bandit applications.
We demonstrated an effective model building process which we used to develop several model extensions that resolve these issues.
To our knowledge, this is the first study illustrating the effects of overdispersion on bandit efficiency and BB-Coint is the first model to indicate finite regret and fast and consistent optional stopping are possible in restless bandits with cointegration.

In the future, we would like to to develop efficient online and more efficient batch versions of the models we presented here, in order to facilitate more exhaustive simulations.
To quantify the advantage gained over time by efficiently stopping and proceeding to the next experiment, we would like to develop an evaluation paradigm with multiple rounds of experimentation and a formal theory of regret in complex bandits with optional stopping.
We would also like to study similar cases in structured bandits, and to explore stopping rules and policies from the $(\epsilon, \delta)-PAC$ and Best Arm Identification literature.

%


\bibliographystyle{IEEEtranS}
\bibliography{IEEEabrv,refs}

\end{document}